\documentclass{article} 
\usepackage{iclr2026_conference,times}

\usepackage{amsmath,amsfonts,bm}

\def\eqref#1{equation~\ref{#1}}

\def\1{\bm{1}}

\DeclareMathAlphabet{\mathsfit}{\encodingdefault}{\sfdefault}{m}{sl}
\SetMathAlphabet{\mathsfit}{bold}{\encodingdefault}{\sfdefault}{bx}{n}

\usepackage{hyperref}
\usepackage{url}                  
\usepackage[utf8]{inputenc}       
\usepackage[T1]{fontenc}          
\usepackage{microtype}            
\usepackage{inconsolata}          
\usepackage{amsmath, amssymb}     
\usepackage{booktabs}             
\usepackage{multirow}
\usepackage{wrapfig}             
\usepackage{capt-of}            
\usepackage{graphicx}
\usepackage[table]{xcolor}       
\usepackage{xspace}
\usepackage{pifont}              
\usepackage{tcolorbox}
\tcbuselibrary{breakable, skins}  
\definecolor{darkblue}{rgb}{0, 0, 0.5}
\definecolor{darkred}{rgb}{0.72, 0.22, 0.27}
\definecolor{lightblue}{RGB}{129, 209, 241}
\definecolor{forestgreen}{RGB}{34, 139, 34}
\definecolor{citepcol}{HTML}{2DDC0E}
\definecolor{tableofcontent}{HTML}{E63E15}
\definecolor{urlcol}{HTML}{2470D8}
\definecolor{myorange}{RGB}{2, 142, 2}
\definecolor{oursbg}{HTML}{EBE0ED}
\definecolor{violet}{RGB}{138, 43, 226}

\hypersetup{
    colorlinks=true,
    citecolor=darkblue,
    linkcolor=red,
    urlcolor=cyan,
    filecolor=magenta,
}
\usepackage{enumitem}             
\newcommand{\imp}[1]{\textcolor{teal}{\textbf{+#1}}}

\newcommand{\xhdr}[1]{{\vspace{1pt}\noindent\bfseries #1}.}
\newcommand{\ie}{\textit{i.e., }}
\newcommand{\eg}{\textit{e.g., }}

\NewDocumentCommand{\shibo}{ mO{} }
{\textcolor{pink}{\textsuperscript{\textit{Shibo}}\textsf{\textbf{\small[#1]}}}}
\newcommand\ours{W2S-OPD\xspace}

\title{Weak-to-Strong On-Policy Distillation}

\author{%
Fangxu Yu\textsuperscript{1}\thanks{Work done during the internship at Microsoft Research.},\;\;
\textbf{Weijia Xu}\textsuperscript{2}, \;\;
\textbf{Michael Xu}\textsuperscript{2}, \;\;
\textbf{Tianyi Zhou}\textsuperscript{\textbf{3}},\;\;
\textbf{Zinan Lin}\textsuperscript{\textbf{2}}
\\
\textsuperscript{1}University of Maryland, College Park,\;\;
\textsuperscript{2}Microsoft Research,\;\;
\textsuperscript{3}MBZUAI\\
}

\newcommand{\msheader}{\textbf{Technical Report}}
\fancypagestyle{msfirst}{%
  \fancyhf{}%
  \fancyhead[L]{\msheader}%
  \fancyhead[R]{}%
  \renewcommand{\headrulewidth}{0.8pt}%
}

\iclrfinalcopy 
\begin{document}

\maketitle
\fancyhf{}%
\fancyhead[L]{\msheader}%
\fancyhead[R]{}%
\renewcommand{\headrulewidth}{0.8pt}%
\thispagestyle{msfirst}

\begin{center}
\vspace{-2.4em}
{ \textbf{Project page:} \url{https://w2s-opd.github.io}}
\end{center}
\vspace{0.3em}

\begin{abstract}
On-policy distillation (OPD), which aligns a student with the teacher's token-level distribution on the student's own rollouts, has become an effective paradigm for transferring capabilities across large language models (LLMs). Prevailing approaches assume a teacher at least as capable as the student, and either distill a larger model into a smaller one, which fails at the frontier when no larger teacher exists, or train multiple domain experts from a shared base and consolidate them into one student, which requires costly training at the student's scale. To tackle these challenges, we introduce Weak-to-Strong On-Policy Distillation (\ours), a simple yet effective OPD framework that improves the strong student by distilling from multiple weak models. Specifically, \ours constructs a proxy teacher in logit space from a contrast pair of a positive and a negative model, both smaller than the student and cheap to obtain.
Their logit difference isolates the capability direction, which is then added to the student's own base model. The resulting proxy teacher thus couples this direction while staying distributionally adjacent to the student. The student then distills it by minimizing the per-token reverse KL on its own rollouts. We instantiate the contrast pair as i) a post-RL expert against its pre-RL initialization, isolating the skill RL instills, ii) a larger against a smaller base model, isolating the capability from scale, and iii) a small base model with correct and wrong hints, isolating the instance-level direction toward the solution. Across four math and three code benchmarks, \ours consistently outperforms OPD and even enables the student to surpass the domain teacher and continues to improve the student when every supervision source is weaker. Further analysis shows that different contrasts yield distinct learning signals: the post-RL and hint contrast emphasizes reasoning frameworks, while the scale contrast emphasizes the solving procedure. 
Our code will be available at \href{https://github.com/Yu-Fangxu/W2S-OPD}{https://github.com/Yu-Fangxu/W2S-OPD}.

\end{abstract}
\section{Introduction}
\begin{figure*}[h]   
\centering
\vspace{-5pt}
\includegraphics[width=1.0\linewidth]{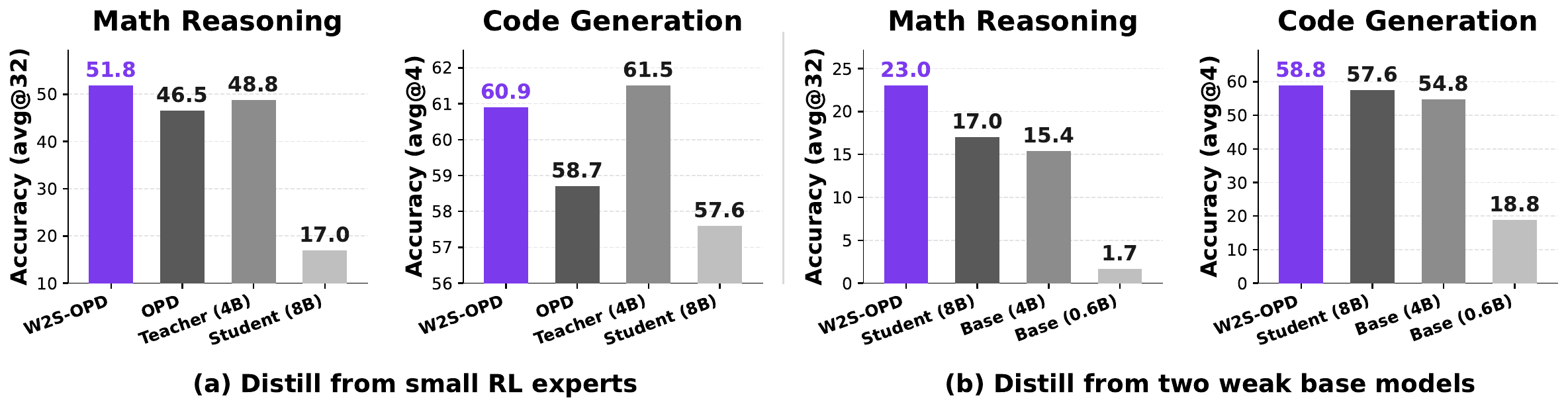}
\vspace{-15pt}
\caption{\textbf{\ours improves Qwen3-8B using 4B models as teachers.} Accuracy is averaged over 4 math reasoning and 3 code generation benchmarks. (a) With a post-RL Qwen3-4B expert, \ours beats OPD. (b) From 2 off-the-shelf base models (Qwen3-4B and 0.6B), both weaker than the student and used without training, \ours still lifts the student above itself and both sources.
}
\label{fig:rubric_analysis}
\end{figure*}

Reinforcement learning with verifiable rewards (RLVR) \citep{guo2025deepseek, shao2024grpo, yu2024flow} and knowledge distillation \citep{gu2024minillm, xiao2026mimo} are two dominant paradigms for improving the reasoning ability of large language models (LLMs). RLVR scales optimization directly from verifiable outcomes, yet its sparse reward provides limited fine-grained supervision. In contrast, knowledge distillation offers dense token-level supervision from a teacher, but the learning from teacher-generated off-policy trajectories suffers from exposure bias. On-policy distillation (OPD) \citep{agarwal2024policy, lu2025onpolicydistillation} combines the complementary strengths of both, which supervises the student with a teacher's token-level fine-grained supervision, yet on trajectories sampled from the student's own policy. This approach delivers dense credit assignment that alleviates exposure bias. However, the premise is a teacher at least as capable as the student.

This premise breaks down in the two ways current practice makes concrete. The strong-to-weak paradigm distills a larger model into a smaller one. For instance, Qwen3~\citep{yang2025qwen3} distills a large-scale model (\eg 235B) into smaller variants (\eg 8B, 14B). As models approach the frontier, however, no larger teacher exists to distill from, capping the performance ceiling. A second line instead performs same-origin consolidation: rather than assuming a single superior teacher, it trains multiple domain experts in parallel from a shared base model and distills their capabilities back into one student. MOPD~\citep{ma2026mopd} instantiates this by applying per-domain RL to obtain a set of domain teachers and merging them in policy space, while DeepSeek-V4~\citep{xu2026deepseek} scales the same specialize-then-consolidate recipe to the trillion-parameter regime. Yet training multiple domain experts at the student's scale is computationally expensive, inflating the cost of OPD. We therefore move to the weak-to-strong paradigm~\citep{burns2023weak, ning2024can}, which induces supervisory signals from weaker models that already exist or can be trained inexpensively. This enables a strong model to keep improving when no stronger teacher is available at a lower cost.

However, directly applying OPD to weak-to-strong learning is challenging. First, the weak teacher's distribution is distinct from the student's, and such a mismatch weakens the distillation signals~\citep{ko2026scaling, li2026rethinking}. Second, since the teacher is weaker than the student, distillation toward it bounds the student at the teacher's level and risks eroding the general capability it already possesses.

To address these challenges, we propose Weak-to-Strong On-Policy Distillation (\ours), a simple yet effective OPD framework that distills the strong student from multiple weak models. Figure~\ref{fig:main_arch} provides an overview. 
Inspired by decoding-time controlled generation~\citep{liu2021dexperts, liu2024tuning}, 
\ours constructs a proxy teacher in logit space from a contrast pair of a positive model and a negative model, both substantially smaller than the student and cheap to obtain. Their logit difference isolates the capability direction, and \ours adds this direction to the student's own base model. The proxy teacher therefore inherits the isolated direction while remaining distributionally close to the student. The student then distills it by minimizing the per-token reverse KL on its own rollouts.

We propose three ways to construct the proxy teacher from weak models: i) \textbf{pre-RL and post-RL models}, where the positive model is a domain expert trained by RL and the negative model is its pre-RL initialization, whose difference isolates the skill that RL instills; since the RL needs only be run at the small model's scale, this transfers an expensive capability to the large student without ever running RL at the student's scale; ii) \textbf{larger and smaller base models}, two off-the-shelf base models of different sizes, whose difference isolates the capability that emerges from scale; this signal comes for free from models that already exist, requiring no new data, reward design, or training; and iii) \textbf{a single base model conditioned on correct and wrong hints}, which extracts the instance-level direction toward the solution from single model and provides token-level supervision more efficiently.
We evaluate all three settings on four math reasoning and three code generation benchmarks.
Our main findings are threefold: 
\begin{itemize}[nosep, leftmargin=1.5em]
\item \textbf{Surpassing the teacher it learns from.} In the pre-RL / post-RL setting, \ours outperforms OPD by 11.4\% and 12.0\% relative on math reasoning under single- and multi-teacher distillation, and lifts the student above the domain expert itself.
\item \textbf{Improving from purely weaker models.} In the smaller / larger and contrastive-hints settings, \ours still improves the student even though every supervision source is weaker than it.
\item \textbf{Different contrasts reinforce complementary reasoning patterns.} The post-RL and hint contrast places more emphasis on the tokens for reasoning framework (\eg planning and monitoring the solving process), whereas the scale contrast emphasizes the solving procedure.
\end{itemize}

More broadly, \ours reframes weak-to-strong learning as isolating and transferring a capability direction rather than imitating a weak supervisor, suggesting how frontier models can keep improving from abundant existing weak signals rather than waiting for a stronger teacher to be built.

\section{Preliminary: On-Policy Distillation}
\label{sec:preliminary}
On-policy distillation (OPD), formalized by the Generalized Knowledge Distillation (GKD) framework \citep{agarwal2024policy}, bridges reinforcement learning and imitation learning by providing dense, token-level supervision on the student's own trajectories. Unlike off-policy distillation that learns from a fixed teacher-generated dataset, OPD lets the student policy $\pi_{S}$ generate rollouts on-policy, and a powerful frozen teacher $\pi_{T}$ then scores every token of these self-generated trajectories, against which the student minimizes a per-token divergence:
\begin{equation}
\label{eq:opd}
\mathcal{L}_{\mathrm{OPD}}(\pi_{S}) \;=\; \mathbb{E}_{x \sim \mathcal{D},\; y \sim \pi_{S}(\cdot \mid x)} \bigg[ \sum_{t=1}^{|y|} D\big( \pi_{S}(\cdot \mid s_t) \,\big\|\, \pi_{T}(\cdot \mid s_t) \big) \bigg],
\end{equation}
where $s_t = (x, y_{<t})$ is the student-generated prefix and $D$ is a per-token divergence. Because states are drawn from the student's own rollouts, the teacher corrects the student precisely on the states the student actually visits, thereby avoiding the exposure bias of off-policy distillation. We instantiate $D$ as the reverse KL divergence $\mathrm{KL}(\pi_{S} \,\|\, \pi_{T})$. In practice, we employ Top-$k$ OPD for computational efficiency; see more details in Appendix~\ref{app:opd_variants}.

Two prominent uses of OPD are strong-to-weak distillation and same-origin consolidation. The former compresses a larger, more capable teacher into a smaller student~\citep{yang2025qwen3}, while the latter distills several RL-trained domain experts back into a single model, with all experts initialized from the student's base so that their distributions stay close~\citep{xiao2026mimo, ma2026mopd}. Both presuppose a teacher at least as capable as the student, which becomes ineffective when training the frontier model, since no larger teacher exists to distill from, and training qualified experts at the student's scale is computationally expensive. These limitations motivate us to instead leverage small, cheaply obtained models to supervise a stronger student.

\section{Weak-to-Strong On-Policy Distillation}
\begin{figure*}[t]
\centering
\includegraphics[width=1.0\textwidth]{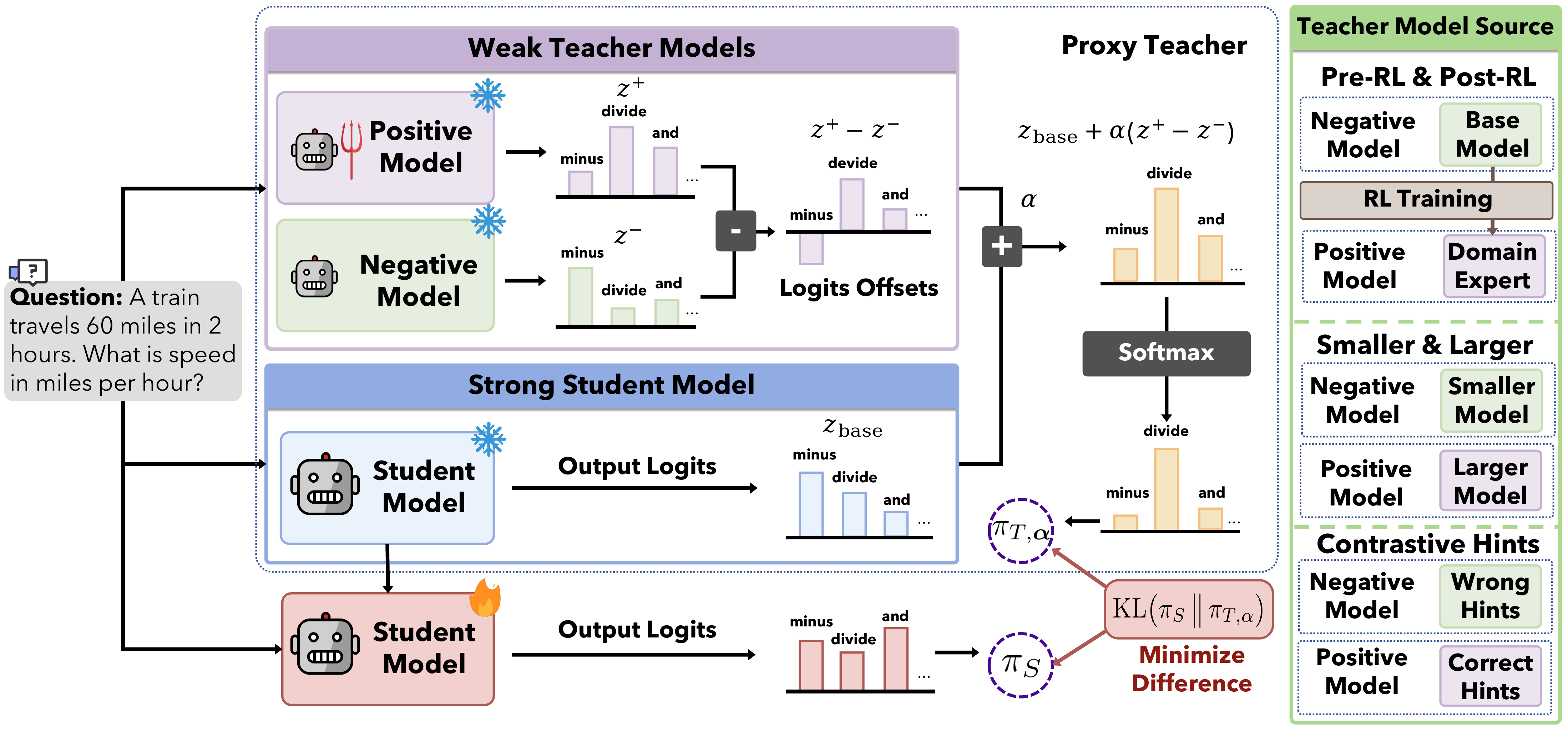}
\caption{Overview of \ours. \ours synthesizes a proxy teacher and distills it into a student. A positive and a negative model form a contrast pair whose logit difference isolates a capability direction. The student $\pi_S$, initialized from the same base model, generates on-policy rollouts and minimizes the per-token reverse KL toward this proxy teacher. The contrast pair can be instantiated as a post-RL expert against its pre-RL initialization, a larger base model against a smaller one, or a single model conditioned on correct against wrong hints.}
\label{fig:main_arch}
\end{figure*}

Improving a strong student with supervision from weaker models presents two central challenges: i) the distributional mismatch between the weak models and the student renders direct distillation ineffective~\citep{ko2026scaling, li2026rethinking}; and ii) since the supervision sources are weaker than the student, direct distillation constrains the student to a lower-capability policy and degrades the general ability it already possesses. To address both challenges, we adopt a directional perspective, as illustrated in Figure~\ref{fig:main_arch}. Instead of imitating a weak model directly, we isolate the capability direction that separates a stronger positive model from a weaker negative model, a signal that is largely disentangled from model scale and hence transferable across scales, and re-anchor this direction onto the student's base model. The resulting proxy teacher is simultaneously equipped with the target domain capability and distributionally adjacent to the student, which stabilizes optimization and preserves the student's general ability. We detail the proxy teacher construction and distillation objective in \S\ref{sec:proxy}, the instantiations of the contrast pair in \S\ref{sec:instantiation}, and its extension to multi-teacher distillation in \S\ref{sec:multi_synthesis}.

\subsection{Proxy Teacher Synthesis and Distillation}
\label{sec:proxy}

Suppose the student $\pi_{S}$ is initialized from a strong base model, while the available weak models form a contrast pair: a positive model $m^{+}$ and a weaker negative model $m^{-}$. Specifically, $m^{+}$ is obtained either by applying domain-specific RL to its initialization $m^{-}$, or taken as a stronger base model than $m^{-}$ (we detail these instantiations in \S\ref{sec:instantiation}). \ours constructs the teacher directly from these three frozen models, requiring only their output logits.

The key observation behind \ours is that, although both $m^{+}$ and $m^{-}$ are weaker than the strong base model, their difference still encodes a transferable capability direction. What the two weak models agree on reflects their shared, limited ability, so subtracting their logits cancels this common component and retains precisely the direction along which $m^{+}$ improves over $m^{-}$. Adding this capability direction onto the strong base model therefore yields a proxy teacher that combines it with the student's general strength while staying distributionally close to the student. We instantiate this construction following decoding-time experts~\citep{liu2021dexperts, liu2024tuning}, where $m^{+}$ acts as the positive model whose logits are additively combined, $m^{-}$ acts as the negative model whose logits are negatively combined, and the student's base model serves as the anchor. Formally, at each position $t$, we condition the three models on the prefix $s_t$ to obtain the logit scores $z_{\mathrm{base}}$, $z^{+}$, and $z^{-}$, and synthesize the proxy teacher as:
\begin{equation}
\label{eq:proxy}
\pi_{T,\alpha}(\cdot \mid s_t) \;=\; \mathrm{softmax}\Big(\, z_{\mathrm{base}}(s_t) \,+\, \alpha \big( z^{+}(s_t) - z^{-}(s_t) \big) \Big),
\end{equation}
where $\alpha \geq 0$ is an amplification coefficient controlling the strength of the injected capability direction, and thus trades off signal strength against distributional proximity.  A small $\alpha$ injects little of the direction and keeps the proxy teacher close to the base model, yielding supervision too weak to drive improvement, whereas a large $\alpha$ injects a stronger signal but risks distorting the distribution and pushing the teacher away from the student. Given the constructed proxy teacher, \ours instantiates the OPD objective in Eq.~\ref{eq:opd} with $\pi_{T} = \pi_{T,\alpha}$:
\begin{equation}
\label{eq:w2sopd}
\mathcal{L}_{\text{W2S-OPD}}(\pi_{S}) \;=\; \mathbb{E}_{x \sim \mathcal{D},\; y \sim \pi_{S}(\cdot \mid x)} \bigg[ \sum_{t=1}^{|y|} \mathrm{KL}\big( \pi_{S}(\cdot \mid s_t) \,\big\|\, \pi_{T,\alpha}(\cdot \mid s_t) \big) \bigg],
\end{equation}
Taken together, \ours address both challenges raised at the beginning of this section. Anchoring the proxy teacher at the student's own base model keeps the supervision target within a distribution the student already realizes (with $\alpha$ bounding how far it moves), and since the weak models enter only through their logit difference rather than their absolute level, distillation transfers their capability direction without pulling the student down toward them, which alleviates the capability ceiling. Appendix~\ref{sec: justification} gives an alternative view of this objective as reward maximization under a KL trust region.

\subsection{Instantiations of the Positive and Negative Models}
\label{sec:instantiation}
\ours requires a contrast pair $(m^{+}, m^{-})$ to construct the proxy teacher. We propose three instantiations and summarize in Table~\ref{tab:instantiations}:

i) \textit{Pre-RL and post-RL}: the positive model $m^{+}$ is a domain expert obtained by applying RL to a small base model, and the negative model $m^{-}$ is its pre-RL initialization, whose difference isolates the domain skill acquired through RL. This is a cheaper way of reducing the cost of training domain experts at the student's own scale.

ii) \textit{Smaller and larger}: the positive and negative models are two
off-the-shelf base models of different sizes (\eg Qwen3-4B and Qwen3-0.6B). Their difference isolates the capability that emerges purely from scale and requires no additional training. This signal is already available in released models and can be used for free to further improve the capability of frontier models.

iii) \textit{Correct and wrong hints}: the positive and negative models are a single base model conditioned on a correct and a wrong solution hint of the identical format, respectively. Since the hint-conditioning is shared, their difference cancels the style shift it induces~\citep{pan2026rlcsd} and isolates the instance-level direction toward the correct solution. Needing only one small model and a reference solution, which token-level supervision efficiently.

These instantiations indicate that \ours can support various origins of the contrast, in which the pair exhibiting a separable capability gap can serve as the supervision source, since the construction of the proxy teacher requires only the output logits of the pair.

\subsection{Unification of Single- and Multi- Proxy Teacher Distillation}
\label{sec:multi_synthesis}
\ours can naturally extend to multiple proxy teacher distillation scenarios. Given $K$ positive models $\{m^{+}_k\}_{k=1}^{K}$, each paired with a weaker negative model $m^{-}_k$, \ours composes their capability directions on the shared base model:
\begin{equation}
\label{eq:multi}
\pi_{T,\boldsymbol{\alpha}}(\cdot \mid s_t) \;=\; \mathrm{softmax}\Big(\, z_{\mathrm{base}}(s_t) \,+\, \sum_{k=1}^{K} \alpha_k \big( z^{+}_k(s_t) - z^{-}_k(s_t) \big) \Big),
\end{equation}
where $\alpha_k$ controls the strength of capability direction $k$. Since each subtraction isolates its own capability direction, the summation injects multiple domain skills into a single proxy teacher, and one distillation run with Eq.~\ref{eq:w2sopd} merges them into a unified student. For example, $\alpha_k$ can be set to $\{0, 1\}$. In this way, each query is routed to its most suitable positive and negative model pair for distillation. 

\begin{table}[t]
\centering
\caption{The three instantiations of the contrast pair in \ours. All models are from Qwen3 series.}
\vspace{5pt}
\resizebox{0.9\linewidth}{!}{
\begin{tabular}{llll}
\toprule
\textbf{Setting} & \textbf{Positive model $m^{+}$} & \textbf{Negative model $m^{-}$} & \textbf{Anchor $z_{\text{base}}$} \\
\midrule
Pre-RL / Post-RL & 4B-RL (post-RL expert) & 4B (pre-RL init)   & 8B (base) \\
Smaller / Larger & 4B (base)              & 0.6B (base)        & 8B (base) \\
Correct / Wrong Hints & 4B w/ correct hint     & 4B w/ wrong hint   & 8B (base) \\
\bottomrule
\end{tabular}%
}
\vspace{-10pt}
\label{tab:instantiations}
\end{table}
\section{Experiments}
We evaluate \ours under the three contrast-pair settings. \S\ref{sec:main_results} reports the main results for each, \S\ref{sec:analysis} examines
the factors behind its effectiveness, and \S\ref{sec:case} characterizes the learning signals the contrast provides.

\subsection{Experimental Setups}
\noindent\textbf{Benchmarks.}
For math reasoning, we evaluate on four competition-level benchmarks: AIME24 \citep{aimo_validation_aime}, AIME25 \citep{aime25}, HMMT25 (Feb.), and HMMT25 (Nov.) \citep{balunovic2025matharena}. For code generation, we adopt HumanEval+, MBPP+ \citep{liu2023your}, and LiveCodeBench-V6 \citep{jain2025livecodebench}. To assess out-of-domain generalization, we further evaluate on GPQA-Diamond~\citep{rein2023gpqa} for scientific reasoning and IFBench~\citep{pyatkin2026generalizing} for instruction following.

\noindent\textbf{Baselines and Evaluation Metrics.}
We compare \ours against OPD, which distills the student directly from the positive model 
under an identical training configuration. As reference points, we additionally report the positive and negative models that form the contrast pair, as well as the student itself. In all evaluations, we set the temperature to 1.0 and top-$p$ to 1.0. On each math reasoning benchmark, we sample 32 solutions per problem, whereas on each code generation benchmark, we sample 4 solutions per problem, and report the average accuracy over all samples on each benchmark for a robust evaluation. We adopt Math-Verify to validate answer correctness for math reasoning.

\noindent\textbf{Training details.}
We use Qwen3-8B \citep{yang2025qwen3} in non-thinking mode as the student $\pi_{S}$; a frozen copy of the same checkpoint serves as the anchor. Both \ours and OPD are trained for 100 steps under an identical configuration. See Appendix~\ref{sec: impl details} for more implementation details.

\noindent\textbf{Contrast pairs.}
Table~\ref{tab:instantiations} summarizes the contrast pair of each setting. In the pre-RL / post-RL setting, the positive model $m^{+}$ is a Qwen3-4B domain expert obtained by applying GRPO \citep{shao2024grpo} to the Qwen3-4B base, for 500 steps on the level-6 subset of DeepMath-103K \citep{he2025deepmath} for math reasoning and for 300 steps on Eurus-RL-Code~\citep{cui2025process} for code generation, which are provided by \citep{yang2026learning}, and the negative model $m^{-}$ is its pre-RL initialization, \ie the original Qwen3-4B model. In the smaller/larger setting, the positive and negative models are the off-the-shelf Qwen3-4B and Qwen3-0.6B base models, requiring no additional training. In the contrastive-hint setting, the positive and negative models share the Qwen3-4B base model, conditioned on a correct and a wrong solution hint of the identical format, respectively.

\subsection{Main Results}
\label{sec:main_results}

\begin{table*}[t]
\centering
\caption{Results for Pre-RL / Post-RL contrast setting. \ours beats OPD and even surpasses the 4B-expert on math. \textit{Improv.} reports the absolute gain over OPD.}
\vspace{5pt}
\resizebox{1.0\textwidth}{!}{
\begin{tabular}{l ccccc cccc}
\toprule
\multirow{2}{*}{\textbf{Method}} & \multicolumn{5}{c}{\textbf{Math Reasoning}} & \multicolumn{4}{c}{\textbf{Code Generation}} \\
\cmidrule(lr){2-6} \cmidrule(lr){7-10}
 & AIME24 & AIME25 & HMMT25 (Feb.) & HMMT25 (Nov.) & Avg. & HumanEval+ & MBPP+ & LCB & Avg. \\
\midrule
Neg. Model (4B)   & 21.6 & 19.9 & 10.0 & 8.2 & 14.9 & 79.9 & 63.6 & 20.9 & 54.8 \\
Pos. Model (4B-RL)  & 62.0 & 57.9 & 34.8 & 40.4 & 48.8 & 86.3 & 71.0 & 27.1 & 61.5 \\
Student Base (8B)  & 26.1 & 21.1 & 11.6 & 9.0  & 17.0 & 81.2 & 70.0 & 21.7 & 57.6 \\
\midrule
\multicolumn{10}{l}{\textit{Single-Teacher Distillation}} \\
SFT & 60.2 & 54.4 & 30.7 & 37.6 & 45.7 & 82.5 & 71.8 & 24.8 & 59.7 \\
OPD      & 62.1 & 54.8 & 30.6 & 38.3 & 46.5 & 83.1 & 71.2 & 22.0 & 58.7 \\\midrule
\rowcolor{oursbg}
\ours    & \textbf{68.9} & \textbf{60.1} & \textbf{35.1} & \textbf{43.1} & \textbf{51.8} & \textbf{84.8} & \textbf{72.4} & \textbf{26.8} & \textbf{60.9} \\
\rowcolor{oursbg}
\textit{Improv.} & \imp{6.8} & \imp{5.3} & \imp{4.5} & \imp{4.8} & \imp{5.3} & \imp{1.7} & \imp{1.2} & \imp{5.0} & \imp{2.2} \\
\midrule
\multicolumn{10}{l}{\textit{Multi-Teacher Distillation}} \\
SFT  & 63.5 & 54.6 & 29.6 & 39.4 & 46.8 & 84.0 & 71.4 & 25.7 & 60.4\\
OPD    & 64.3 & 54.9 & 29.6 & 37.4 & 46.5 & 83.4 & 69.5 & 25.6 & 59.5 \\\midrule
\rowcolor{oursbg}
\ours    & \textbf{67.4} & \textbf{61.0} & \textbf{34.7} & \textbf{45.2} & \textbf{52.1} & \textbf{85.2} & \textbf{71.0} & \textbf{27.1} & \textbf{61.1} \\
\rowcolor{oursbg}
\textit{Improv.} & \imp{3.1} & \imp{6.1} & \imp{5.1} & \imp{7.8} & \imp{5.6} & \imp{1.8} & \imp{1.5} & \imp{1.5} & \imp{1.6} \\
\bottomrule
\end{tabular}
}

\label{tab:main}
\vspace{-10pt}
\end{table*}

\begin{table*}[t]
\centering
\caption{Results for Smaller and Larger contrast setting.  \ours improves the 8B student above its own base even though both source models are weaker than it. \textit{Improv.} reports the absolute gain of \ours{} over the student base model.}
\vspace{5pt}
\resizebox{\textwidth}{!}{
\begin{tabular}{l ccccc cccc}
\toprule
\multirow{2}{*}{\textbf{Method}} & \multicolumn{5}{c}{\textbf{Math Reasoning}} & \multicolumn{4}{c}{\textbf{Code Generation}} \\
\cmidrule(lr){2-6} \cmidrule(lr){7-10}
 & AIME24 & AIME25 & HMMT25 (Feb.) & HMMT25 (Nov.) & Avg. & HumanEval+ & MBPP+ & LCB & Avg. \\
\midrule
Neg. Model (0.6B) & 1.6 & 2.3 & 0.1 & 2.7 & 1.7 & 25.6 & 26.3 & 4.6 & 18.8 \\
Pos. Model (4B)   & 21.6 & 19.9 & 10.0 & 8.2 & 14.9 & 79.9 & 63.6 & 20.9 & 54.8 \\
Student Base (8B)  & 26.1 & 21.1 & 11.6 & 9.0  & 17.0 & 81.2 & 70.0 & 21.7 & 57.6 \\
\midrule
\rowcolor{oursbg}
\ours & \textbf{33.1} & \textbf{25.9} & \textbf{13.0} & \textbf{20.1} & \textbf{23.0} & \textbf{81.7} & \textbf{71.1} & \textbf{23.7} & \textbf{58.8} \\
\rowcolor{oursbg}
\textit{Improv.} & \imp{7.0} & \imp{4.8} & \imp{1.4} & \imp{11.1} & \imp{6.0} & \imp{0.5} & \imp{1.1} & \imp{2.0} & \imp{1.2} \\
\bottomrule
\end{tabular}
}

\label{tab:two_weak}
\vspace{-10pt}
\end{table*}

\begin{figure*}[t]
\centering
\includegraphics[width=1.0\textwidth]{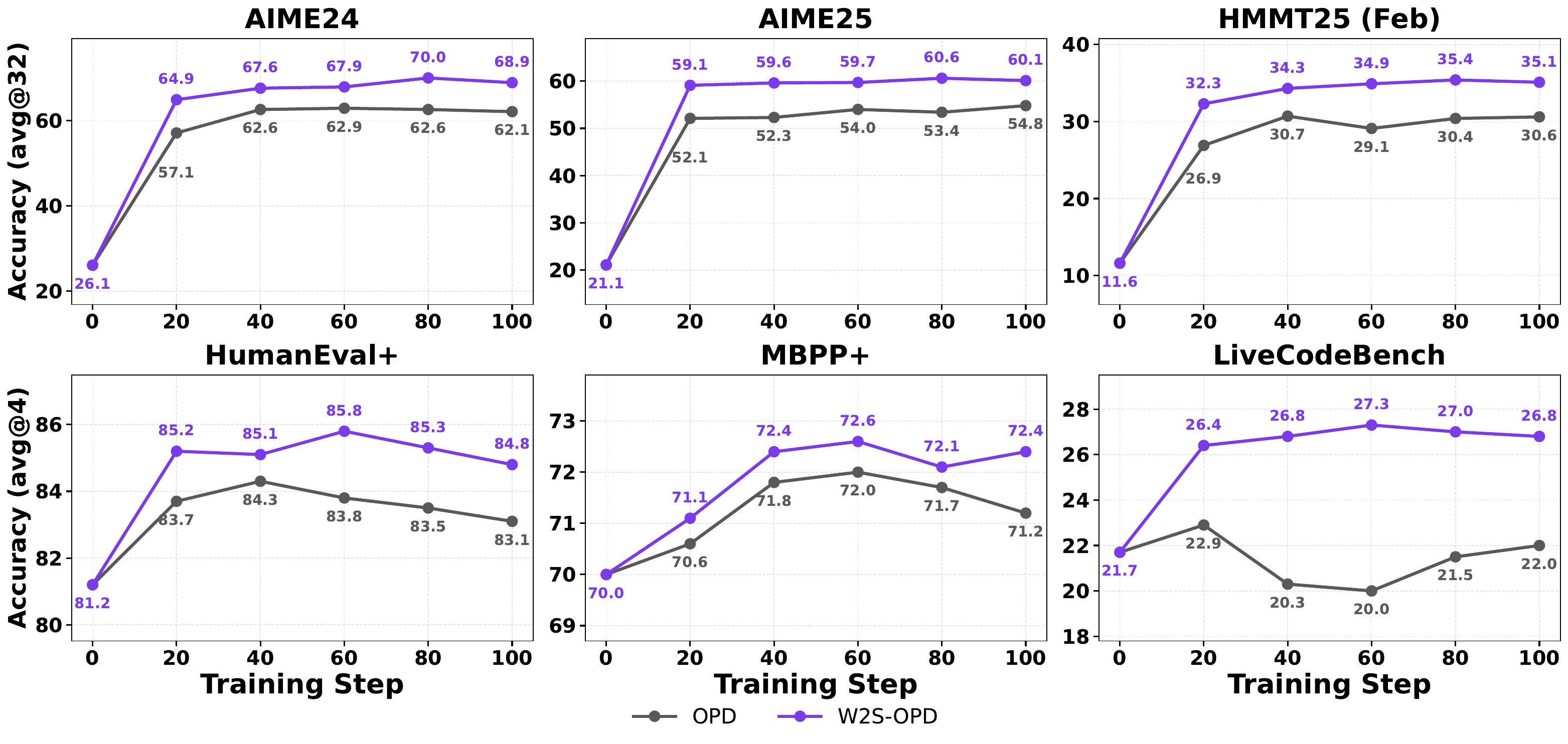}
\caption{Performance on the math and code benchmarks over training steps. \ours improves faster and outperforms OPD.}
\label{fig:step}
\vspace{-10pt}
\end{figure*}

\textbf{i) Distillation from Pre-RL / Post-RL.}
Table~\ref{tab:main} compares \ours against OPD under single-teacher distillation, which learns from a single domain expert, and multi-teacher distillation, which merges multiple domain experts (here a math and a code expert) into one student. We have the following observations:
\noindent\textbf{(a) A small post-RL expert substantially improves the large student.} \ours consistently outperforms SFT and OPD across all benchmarks in the single-teacher setting, with 11.4\% and 3.7\% average relative improvements on math and code, respectively. Notably, on math reasoning \ours surpasses the domain teacher itself, whereas OPD remains below it. On code generation, \ours likewise narrows the gap to the teacher. \textbf{(b) \ours outperforms OPD when distilling from multiple domain teachers.} We conduct experiments in the multi-teacher setting, where we aim to merge the capabilities from different domain teachers into a single student through OPD, which is obtained by applying domain-specific RL to the same base model. Specifically, we mix the math and coding training data and route each sample to the corresponding domain teacher through their domain labels. The results in Table~\ref{tab:main} demonstrate that \ours consistently leads to better performance than OPD on all benchmarks. 
\textbf{(c) \ours learns faster and more stably.} Figure~\ref{fig:step} tracks benchmark performance over training and shows that \ours outperforms OPD throughout the training.

\begin{table*}[t]
\centering
\caption{Results for contrastive hints setting.  \ours improves the 8B student above its own base with only a single weaker and smaller model. \textit{Improv.} reports the absolute gain of \ours{} over the student base model.
}
\vspace{5pt}
\resizebox{\textwidth}{!}{
\begin{tabular}{l ccccc cccc}
\toprule
\multirow{2}{*}{\textbf{Method}} & \multicolumn{5}{c}{\textbf{Math Reasoning}} & \multicolumn{4}{c}{\textbf{Code Generation}} \\
\cmidrule(lr){2-6} \cmidrule(lr){7-10}
 & AIME24 & AIME25 & HMMT25 (Feb.) & HMMT25 (Nov.) & Avg. & HumanEval+ & MBPP+ & LCB & Avg. \\
\midrule
Pos./Neg. Model (4B) & 21.6 & 19.9 & 10.0 & 8.2 & 14.9 & 79.9 & 63.6 & 20.9 & 54.8 \\
Student Base (8B)  & 26.1 & 21.1 & 11.6 & 9.0  & 17.0 & 81.2 & 70.0 & 21.7 & 57.6 \\
\midrule
\rowcolor{oursbg}
\ours & \textbf{27.7} & \textbf{23.3} & \textbf{12.0} & \textbf{10.7} & \textbf{18.4} & \textbf{82.5} & \textbf{70.8} & \textbf{22.7} & \textbf{58.7}\\
\rowcolor{oursbg}
\textit{Improv.} & \imp{1.6} & \imp{2.2} & \imp{0.4} & \imp{1.7} & \imp{1.4} & \imp{1.3} & \imp{0.8} & \imp{1.0} & \imp{1.1} \\
\bottomrule
\end{tabular}
}
\label{tab:hints}
\vspace{-5pt}
\end{table*}

\noindent\textbf{ii) Distillation from Smaller / Larger Base Models.}
In this setting, we find that \textbf{Two weak base models can improve the stronger student.} As shown in Table~\ref{tab:two_weak}, when the proxy teacher is constructed from Qwen3-4B and Qwen3-0.6B, both weaker than the student, \ours still improves the
student by an absolute 6.0\% on math reasoning and 1.2\% on code generation on average. This confirms that the inherent gap between two off-the-shelf weak models also encodes a transferable improving direction to distill from.

\noindent\textbf{iii) Distillation from a Single Model with Correct / Wrong Hints.}
As shown in Table~\ref{tab:hints}, a contrastive hint direction from a single weak model improves the student. \ours improves the student by an absolute 1.4\% on math reasoning and 1.1\% on code generation on average, despite the hint model being a 4B model weaker than the student. This construction distills the privileged information and shows that a meaningful learning signal can also be extracted from a difference in context without requiring two different models.

\subsection{Training Analysis}
\label{sec:analysis}
Using the Pre-RL / Post-RL setting, we conduct two further analyses: the impact of the amplification coefficient $\alpha$ on distillation, and out-of-domain generalization compared with OPD.
\subsubsection{Effect of the Amplification Coefficient $\alpha$}
\label{sec:alpha}

\begin{figure*}[t]
\centering
\includegraphics[width=1.0\textwidth]{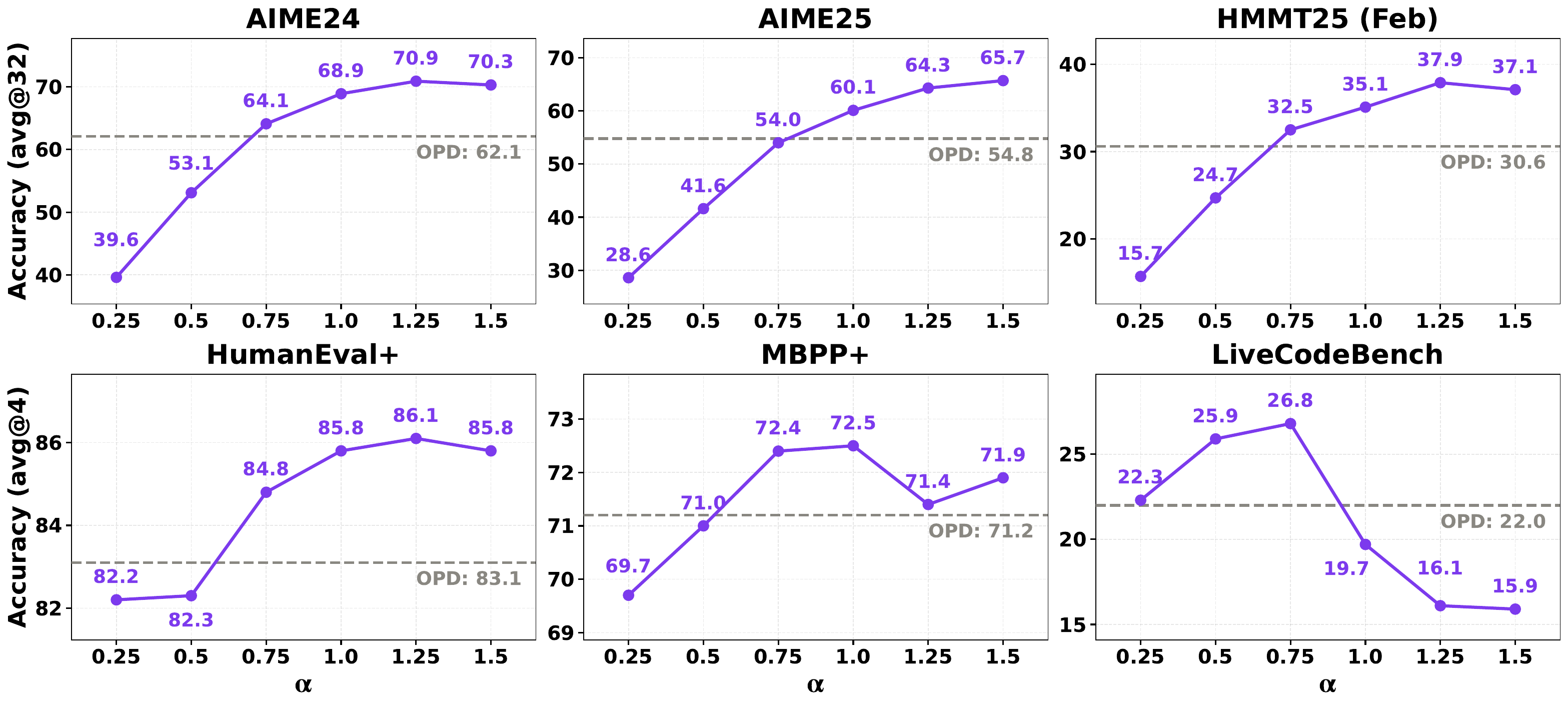}
\caption{Performance on the math and code benchmarks with different $\alpha$. OPD is included for reference, denoted by the gray line. \ours outperforms OPD over a wide range of $\alpha$. }
\label{fig:alpha}
\vspace{-10pt}
\end{figure*}

To investigate the impact of the amplification coefficient, we vary $\alpha$ with all other configurations fixed and report the results in Figure~\ref{fig:alpha}. When $\alpha$ is small, $\pi_{T,\alpha}$ stays close to the student's own base model, where the capability direction is barely injected, the distillation signal is weak, and the distilled student falls behind OPD. As $\alpha$ increases, more domain skill is injected into the proxy teacher, and the student's performance improves accordingly. When the $\alpha$ continues increasing, performance starts to decline on most benchmarks, likely because an overly large $\alpha$ drives the proxy teacher further away from the student's distribution, making the supervision increasingly hard for the student to learn. Consequently, a moderate $\alpha$ strikes the best balance between the strength of the injected signal and the adjacency of the teacher.

\subsubsection{Generalization to Out-of-Domain Tasks}
\label{sec:ood}
\begin{wraptable}{r}{0.52\textwidth}
\centering

\vspace{-20pt}
\caption{Results for OOD generalization on GPQA-Diamond and IFBench. Both distillation methods are trained only on the math task, \ours transfers out of domain and improves general ability, whereas OPD can degrade it below the base. \textit{Improv.} indicates the absolute gain over OPD.}
\vspace{5pt}
\begin{tabular}{l c c}
\toprule
\textbf{Method} & \textbf{GPQA-Diamond} & \textbf{IFBench}\\
\midrule
Positive Model  & 51.8 & 26.0\\
Student Base  & 38.9 & 26.3\\
\midrule
OPD                        & 54.4 & 25.9\\
\rowcolor{oursbg}
\ours                      & \textbf{56.5} & 27.0 \\
\rowcolor{oursbg}
\textit{Improv.}           & \imp{2.1} & \imp{1.1}\\
\bottomrule
\end{tabular}
\label{tab:gpqa}
\end{wraptable}
A crucial concern for weak-to-strong distillation is whether learning from a small domain expert erodes the strong student's general capability. To investigate this, we evaluate the student trained in the Pre-RL / Post-RL setting on two out-of-domain benchmarks. As shown in Table~\ref{tab:gpqa}, the distilled reasoning skill transfers well beyond the training domain. \ours lifts the student from 38.9 to 56.5 on GPQA-Diamond, outperforming OPD by an absolute 2.1 \%. The two models diverge on IFBench, where OPD degrades the student below its initialization, consistent with the student absorbing the small expert's limitations, yet \ours instead improves compared to the base model and outperforms OPD by 1.1\%. These results indicate that \ours transfers domain skill without sacrificing and even enhancing the student's general ability.

\subsection{What Kinds of Tokens Are Most Strengthened?}
\label{sec:case}
Since all settings improve the student, we raise the question of what each capability direction actually provides. We quantitatively analyze this question by recording each token's offset between the positive and negative models on a shared solution generated by the student model. At every position $t$ along this trace, the capability direction in logit space is
$\Delta z_t = z^{+}(s_t) - z^{-}(s_t)$. To measure how strongly it reinforces the realized token $x_t$, we take the log-probability each model assigns to $x_t$ and define:
\begin{equation}
\label{eq:delta}
\Delta_t \;=\; \log \pi^{+}(x_t \mid s_t) \;-\; \log \pi^{-}(x_t \mid s_t),
\qquad \pi^{\pm}=\mathrm{softmax}(z^{\pm}).
\end{equation}

Tokens with large $\Delta_t$ are those most strongly reinforced by the
capability direction. To identify which reasoning steps each contrast reinforces, we collect the top-1\% highest-$\Delta$ tokens across 200 math-reasoning traces and label every token with one of eight problem-solving episodes~\citep{schoenfeld2014mathematical} using ThinkARM~\citep{li2026schoenfeld}, an automatic episode classifier. Table~\ref{tab:episode-share} shows the token distribution. Relative to the scale contrast, the post-RL and hint contrasts place more weight on the reasoning \emph{framework} such as the \textit{Plan} and \textit{Monitor} episodes that structure and track the solution, whereas the scale contrast keeps more weight on the core solving steps, \textit{Analyze} and \textit{Implement}. The hint contrast additionally concentrates on the final \textit{Answer} tokens, which is a consequence of conditioning on a correct versus a wrong answer.
Figure~\ref{fig:case} illustrates the same division on a single trace: \textbf{the three contrasts reinforce different kinds of tokens.} These emphases are complementary, resulting in a student learning to improve the reasoning by distilling different patterns.

\begin{table}[t]
\centering
\caption{Distribution of the top-1\% highest-$\Delta$ tokens over the eight Schoenfeld episodes. The first
row gives the distribution of each episodes of all tokens.}
\vspace{5pt}
\label{tab:episode-share}
\resizebox{\textwidth}{!}{
\begin{tabular}{lrrrrrrrr}
\toprule
 & Read & Analyze & Plan & Implement & Explore & Verify & Monitor & Answer \\
\midrule
\textit{All tokens} & 10.3 & 33.2 & 8.6 & 40.7 & 0.5 & 2.4 & 0.5 & 3.8 \\
\midrule
Pre-RL / Post-RL & 9.6 & 40.5 & 15.7 & 24.0 & 1.2 & 4.3 & 1.8 & 2.9 \\
Smaller / Larger  & 6.5 & 44.4 & 12.6 & 27.7 & 0.8 & 4.2 & 0.9 & 2.9 \\
Correct / Wrong Hints & 5.2 & 29.8 & 20.2 & 19.6 & 1.1 & 5.4 & 2.3 & 16.4 \\
\bottomrule
\end{tabular}
}

\end{table}
\providecommand{\hlstrut}{\rule[-0.30ex]{0pt}{2.05ex}}
\providecommand{\hlbox}[2]{\raisebox{0pt}[0pt][0pt]{\setlength{\fboxsep}{0pt}%
  \colorbox{#1}{\hlstrut #2}}}
\providecommand{\hlR}[2]{\hlbox{red!#1}{#2}} 
\providecommand{\hlB}[2]{\hlbox{blue!#1}{#2}}  
\providecommand{\hlG}[2]{\hlbox{teal!#1}{#2}}   
\providecommand{\gry}[1]{\textcolor{black!45}{#1}}
\providecommand{\panelw}{0.325\textwidth}
\providecommand{\panelh}{34ex}

\begin{figure*}[t]
\centering
\small
\fbox{\parbox{0.965\textwidth}{\textbf{Question:} Calculate the line integral of $\frac{1}{z}$ over a contour that consists of a square and a circle, both centered at the origin, oriented counterclockwise. \hfill \textit{(answer $4\pi i$; wrong-answer hint $2\pi i$)}}}
\\[6pt]
\begin{minipage}[t]{\panelw}\centering
\fbox{\parbox[t]{0.90\linewidth}{\raggedright\linespread{1.3}\selectfont
\gry{We are} \hlR{35}{integrating} \gry{$\oint_C \frac{1}{z}dz$ $\cdots$ Step 1: Understand the function $\cdots$ $1/z$ has a} \hlR{39}{singularity} \gry{at $z{=}0$, inside both $\cdots$} \hlR{45}{Step 2: Cauchy's Theorem} \gry{$\cdots$} \hlR{48}{However}\gry{, it applies only to holomorphic $f$ $\cdots$} \hlR{55}{But} \gry{we can use the Integral} \hlR{41}{Formula} \gry{$\cdots$} \hlR{35}{Step 3: known result} \gry{$\cdots$ positively} \hlR{45}{oriented} \gry{$\Rightarrow 2\pi i$ $\cdots$ so the total $=2\pi i{+}2\pi i=4\pi i$ $\cdots$ Final Answer: $4\pi i$.}
}}\\[3pt]
{\footnotesize (a) Pre-RL / Post-RL}
\end{minipage}\hfill
\begin{minipage}[t]{\panelw}\centering
\fbox{\parbox[t]{0.90\linewidth}{\raggedright\linespread{1.3}\selectfont
\gry{We are integrating $\oint_C \frac{1}{z}dz$ $\cdots$ Step 1: Understand the} \hlB{46}{function} \gry{$\cdots$ $1/z$ has a singularity at $z{=}0$, inside} \hlB{48}{both} \gry{$\cdots$ \\Step 2: Cauchy's Theorem $\cdots$} \gry{However, it applies only to} \hlB{23}{holomorphic $f$} \gry{$\cdots$ But we can} \hlB{25}{use} \gry{the Integral Formula $\cdots$ Step 3: known result} \gry{$\cdots$} \hlB{60}{positively oriented} \gry{$\Rightarrow 2\pi i$ $\cdots$ so the} \hlB{32}{total} \gry{$=$}\hlB{35}{$2\pi i{+}2\pi i$}\gry{$=4\pi i$ $\cdots$ Final Answer: $4\pi i$.}
}}\\[3pt]
{\footnotesize (b) Smaller / Larger Models}
\end{minipage}\hfill
\begin{minipage}[t]{\panelw}\centering
\fbox{\parbox[t]{0.90\linewidth}{\raggedright\linespread{1.3}\selectfont
\gry{We are integrating $\oint_C \frac{1}{z}dz$ $\cdots$ Step 1: Understand the function $\cdots$ $1/z$ has a singularity at $z{=}0$, inside both $\cdots$} \hlG{10}{Step 2: Cauchy's Theorem} \gry{$\cdots$ However, it applies only to holomorphic $f$ $\cdots$ But we can use the Integral Formula $\cdots$ Step 3: known result $\cdots$ positively oriented $\Rightarrow 2\pi i$ $\cdots$ so the} \hlG{12}{total} \gry{$=2\pi i{+}2\pi i=$}\hlG{60}{$4$}\gry{$\pi i$ $\cdots$} \hlG{18}{Final Answer}\gry{: $4\pi i$.}
}}\\[3pt]
{\footnotesize (c) Correct / Wrong Hints }
\end{minipage}
\caption{A case study that the three contrasts strengthen different reasoning tokens. The shade of color represents the relative magnitude of $\Delta_t$.}
\label{fig:case}
\vspace{-10pt}
\end{figure*}

\section{Related Work}
\textbf{On-Policy Distillation.} OPD~\citep{lu2025onpolicydistillation, song2026survey, agarwal2024policy} supervises the student on its own rollouts with a superior teacher. It has become a key post-training paradigm for strong-to-weak distillation~\citep{zeng2026glm, yang2025qwen3} and for merging multi-domain experts into one model~\citep{xiao2026mimo, xu2026deepseek, chen2026counteraction, yang2026oprd}, with follow-ups relaxing its requirements through self-distillation from privileged information~\citep{zhao2026self, shenfeld2026self, hubotter2026reinforcement, ye2026policy}, stabilized optimization against the student--teacher gap~\citep{jin2026entropy, ko2026scaling, xing2026trust, jang2026stable}, and multimodal extensions~\citep{li2026video, liu2026visual, yoon2026decomposed, yuan2026vision, liang2026visual}.
These works assume a teacher at least as capable as the student. In contrast, \ours works weak-to-strong, synthesizing a teacher from weak models rather than requiring a strong superior model. 
Direct-OPD~\citep{feng2026weak} is a concurrent work that transfers a weak teacher's pre-/post-RL log-ratio as a dense reward. However, its contrast is confined to RL-trained teachers, without incorporating directions from multiple contrastive sources, and is only evaluated on math reasoning tasks.

\textbf{Weak-to-Strong Generalization.}
Weak-to-Strong elicits the capabilities of a stronger model with the supervision of a weak model~\citep{burns2023weak}. This is critical when a stronger model is hard to obtain~\citep{christiano2018supervising}. Recent work extends it to LLM reasoning and alignment~\citep{yao2025revisiting, yao2025weak, yuan2026incentivizing, yang2024weak, zhao2024weak}. Another line casts learning by teaching, where a strong model improves by instructing weaker students and turning their comprehension into a training signal~\citep{ning2024can, cetin2026reinforcement}.
However, such supervision primarily provides trajectory-level signals, while \ours instead distills a synthesized proxy teacher with token-level dense supervision.

\textbf{Controllable Text Generation.} Controlling the outputs of an LLM has been widely studied~\citep{prabhumoye2020exploring} along two directions: training-based methods finetune the model to elicit desired properties~\citep{keskar2019ctrl, chan2020cocon} but are computationally costly, whereas decoding-time methods steer generation by shifting or composing output distributions at inference~\citep{liu2021dexperts, krause2021gedi, qin2022cold}. \ours adopts this decoding-time composition to synthesize its proxy teacher, but turns it into a training signal that transfers the ability from weak models to a strong student.

\section{Conclusion}
We introduce \ours, a weak-to-strong on-policy distillation framework that improves a strong student using only smaller or weaker models. By isolating the capability direction between a contrast pair of weak models and re-anchoring it onto the student's base model, \ours synthesizes a proxy teacher that couples the isolated capability while staying distributionally adjacent to the student. Across math reasoning and code generation benchmarks, \ours substantially outperforms OPD, surpasses the weak teacher itself, and remains effective when the contrast pair comes from RL training, model scale, or contrastive hints, while preserving out-of-domain capabilities.

\textbf{Discussion.} As models approach the frontier, the assumption of an ever-stronger teacher no longer holds, and where the supervision for OPD should come from becomes an open question. Weak supervision sources are abundant, yet how far they can push a stronger student before weak-to-strong supervision saturates, and how to elicit more informative signals from them, remain to be explored. Future work may develop weak-to-strong learning into a sustained paradigm for OPD and post-training more broadly, in which frontier models continue to improve from supervision sources cheaper and weaker than themselves.

\bibliography{iclr2026_conference}
\bibliographystyle{iclr2026_conference}
\appendix
\newpage
\section{KL-Divergence Estimation}
\label{app:opd_variants}

\xhdr{Single-token OPD}
An alternative estimator, adopted by several recent OPD implementations~\citep{lu2025onpolicydistillation, zhu2026hybrid}, extracts a single scalar of teacher information per sampled token: the log-ratio on the sampled token is treated as a token-level advantage,
\begin{equation}
\label{eq:single_token}
A_t \;=\; \log \pi_{T}(y_t \mid s_t) \;-\; \log \pi_{S}^{\mathrm{old}}(y_t \mid s_t),
\end{equation}
and plugged into a clipped policy-gradient update, where $\pi_{S}^{\mathrm{old}}$ denotes the policy that generated the rollouts. This variant reuses the RL infrastructure, but conveys only 1 scalar per position and thus exhibits higher variance than the dense top-$K$ objective, which backpropagates the teacher's distribution at every position through a directly differentiable divergence and requires no advantage estimation or clipping. We therefore adopt the dense top-$K$ objective for both distillation arms in all experiments.

\xhdr{Top-$K$ OPD}
Materializing the teacher's full distribution requires storing and transferring $|\mathcal{V}|$ logits (roughly 150K for the Qwen3 family) at every response position, which is prohibitively expensive to communicate across training workers. Since the teacher distribution is highly peaked, its top-$K$ tokens already carry the majority of probability mass at nearly all positions. The teacher worker therefore ships only the top-$K$ token indices and log-probabilities per position, and the divergence in Eq.~\ref{eq:opd} is estimated on this truncated support:
\begin{equation}
\label{eq:topk}
\mathcal{L}^{\text{top-}K}_{\mathrm{OPD}}(\pi_{S}) \;=\; \mathbb{E}_{x \sim \mathcal{D},\; y \sim \pi_{S}(\cdot \mid x)} \bigg[ \sum_{t=1}^{|y|} \mathrm{KL}\big( \bar{\pi}_{S}^{\mathcal{K}_t} \,\big\|\, \bar{\pi}_{T}^{\mathcal{K}_t} \big) \bigg],
\end{equation}
where $\mathcal{K}_t$ denotes the top-$K$ support of the teacher distribution at position $t$, and $\bar{\pi}^{\mathcal{K}_t}$ denotes the corresponding distribution restricted to $\mathcal{K}_t$ and renormalized. The truncation error is bounded by the tiny probability mass that falls outside the teacher's top-$K$ set, so the estimation preserves the supervision almost losslessly while reducing the cost.

\section{From logit arithmetic to a KL-constrained objective}
\label{sec: justification}
Let $r(s,a):=\log\pi^{+}(a\mid s)-\log\pi^{-}(a\mid s)$ denote the \emph{capability
reward} carried by the contrast pair: a token-level, relative quantity measuring how
much more the positive model favors $a$ than the negative one does, in which the
absolute competence of either source never appears. Since $e^{z^{+}(a)-z^{-}(a)}$
equals $\pi^{+}(a\mid s)/\pi^{-}(a\mid s)$ up to a factor independent of $a$, which
the softmax normalizer absorbs, Eq.~\ref{eq:proxy} is an exponential tilting of the
student's own base model,
\begin{equation}
  \pi_{T,\alpha}(a\mid s)=\frac{\pi_{\mathrm{base}}(a\mid s)\,e^{\alpha r(s,a)}}{Z_{\alpha}(s)},
  \qquad
  Z_{\alpha}(s)=\mathbb{E}_{a\sim\pi_{\mathrm{base}}(\cdot\mid s)}\big[e^{\alpha r(s,a)}\big].
  \label{eq:tilt}
\end{equation}
Substituting Eq.~\ref{eq:tilt} into $\mathrm{KL}(q\,\|\,\pi_{T,\alpha})$ and
rearranging gives, for every $q\in\Delta(\mathcal{V})$,
\begin{equation}
  \underbrace{\mathbb{E}_{a\sim q}\big[r(s,a)\big]
  -\tfrac{1}{\alpha}\mathrm{KL}\big(q\,\|\,\pi_{\mathrm{base}}(\cdot\mid s)\big)}_{\textstyle \mathcal{J}_{\alpha}(q\,;s)}
  \;=\;\tfrac{1}{\alpha}\log Z_{\alpha}(s)\;-\;\tfrac{1}{\alpha}\mathrm{KL}\big(q\,\|\,\pi_{T,\alpha}(\cdot\mid s)\big).
  \label{eq:variational}
\end{equation}
The first term on the right does not depend on $q$ and the second is nonnegative and
vanishes only at $q=\pi_{T,\alpha}$. The proxy teacher is therefore the unique
closed-form maximizer
\begin{equation}
  \pi_{T,\alpha}(\cdot\mid s)\;=\;\arg\max_{q\in\Delta(\mathcal{V})}
  \Big\{\underbrace{\mathbb{E}_{a\sim q}\big[r(s,a)\big]}_{\text{inject capability}}
  \;-\;\tfrac{1}{\alpha}\underbrace{\mathrm{KL}\big(q\,\|\,\pi_{\mathrm{base}}(\cdot\mid s)\big)}_{\text{stay adjacent to the student}}\Big\},
  \label{eq:argmax}
\end{equation}
where $1/\alpha$ is the strength of a trust region centered at
the student rather than at any weak model.

\noindent\textbf{A compositional view.}
Eq.~\ref{eq:tilt} can be stated without reference to logits at all,
\begin{equation}
  \pi_{T,\alpha}(a \mid s)\;\propto\;\pi_{\mathrm{base}}(a \mid s)\,
  \Big(\frac{\pi^{+}(a\mid s)}{\pi^{-}(a\mid s)}\Big)^{\alpha},
  \label{eq:poe}
\end{equation}
which exhibits the proxy teacher as a product of experts over the shared vocabulary:
the student's base model is the anchor, $\pi^{+}$ enters as a conjoined expert and
$\pi^{-}$ as a negated one, and $\alpha$ is the exponent controlling how sharply the
pair is applied. The capability direction is thus a \emph{negation} operation, which divides out what the two weak models agree on and the trust region in Eq.~\ref{eq:argmax} is precisely the price of keeping the resulting product near the
anchor, with $\tfrac{1}{\alpha}\log Z_{\alpha}(s)$ its free energy at $s$. With $K$
contrast pairs the same reading gives:
\begin{equation}
  \pi_{T,\boldsymbol{\alpha}}(a \mid s)\;\propto\;\pi_{\mathrm{base}}(a\mid s)
  \prod_{k=1}^{K}\Big(\frac{\pi^{+}_{k}(a \mid s)}{\pi^{-}_{k}(a\mid s)}\Big)^{\alpha_{k}},
  \label{eq:multipoe}
\end{equation}
so merging capability directions is multiplication of experts. In particular the
result is invariant to the order in which the pairs are introduced, unlike
sequentially fine-tuning or distilling one teacher after another, where the outcome
depends on the schedule and earlier skills may be overwritten.

Three design choices follow: i) \emph{Adjacency}: the reference measure in
Eq.~\ref{eq:argmax} is the student's own base model, so the proxy teacher is a
reweighting of a distribution the student already realizes, and $\alpha$ sets how far
it may move; $\alpha\!\to\!0$ recovers $\pi_{\mathrm{base}}$ and supplies no
supervision, whereas $\alpha\!\to\!\infty$ removes the trust region and collapses onto
$\arg\max_{a} r(s,a)$, so a moderate $\alpha$ is optimal (Figure~\ref{fig:alpha}).
ii) \emph{Contrast}: the weak models enter only through the ratio $r$, so a pair with no
capability gap leaves the anchor untouched and the student is never ceilinged at the
sources, unlike direct OPD whose optimum is $\pi^{+}$ itself. iii) \emph{Composition}:
rewards add where policies do not, so Eq.~\ref{eq:multi} is the same maximizer with
composite reward $\sum_{k}\alpha_{k}r_{k}$, and $K$ capability directions are merged
by a single distillation run.

\section{Implementation Details}
All experiments are implemented on top of the verl framework, and each distillation run is conducted on 2 NVIDIA B200 GPUs. Table~\ref{tab:hyperparameters} lists the training hyperparameters shared by \ours and OPD. For the proxy teacher, the divergence is estimated on the teacher's top-$K$ support with $K=32$ (Appendix~\ref{app:opd_variants}). Table~\ref{tab:expert_hparams} reports the GRPO recipe used to obtain domain experts in the pre-RL / post-RL setting, and Table~\ref{tab:train_prompt} shows the rollout instructions used during OPD training.
\label{sec: impl details}

\begin{table}[t]
\centering
\setlength{\tabcolsep}{4pt}
\begin{minipage}[t]{0.48\textwidth}
\vspace{5pt}
\centering
\scriptsize
\caption{Training hyperparameters of \ours{}. Paired entries denote math / code.}
\label{tab:hyperparameters}
\begin{tabular}{ll}
\toprule
\textbf{Hyperparameter} & \textbf{Value} \\
\midrule
Algorithm & On-Policy Distillation \\
Student Model & Qwen3-8B \\
Positive / Negative & Qwen3-4B-RL / Qwen3-4B \\
Amplification Coeff. ($\alpha$) & 1.0 / 0.75 \\
Teacher Top-$K$ & 32 \\
Max Prompt Length & 2048 \\
Max Response Length & 16384 \\
Learning Rate & 2e-6 \\
LR Scheduler & Constant \\
Sampling Temp. / Top-$p$ & 1.0 / 1.0 \\
Batch Size & 64 \\
Training Steps & 100 \\
\bottomrule
\end{tabular}

\end{minipage}
\hfill
\begin{minipage}[t]{0.48\textwidth}
\vspace{5pt}
\centering
\scriptsize
\caption{Training hyperparameters of the domain experts (RL-trained positive models).}
\label{tab:expert_hparams}
\begin{tabular}{ll}
\toprule
\textbf{Hyperparameter} & \textbf{Value} \\
\midrule
Algorithm & GRPO \\
Base Models & Qwen3-4B \\
Reasoning Mode & Non-Thinking \\
Max Prompt Length & 2048 \\
Max Response Length & 16384 \\
KL Loss Coefficient & 0 \\
Learning Rate & 1e-6 \\
LR Scheduler & Constant \\
Sampling Temp. / Top-$p$ & 1.0 / 1.0 \\
Batch Size & 128 \\
Mini-batch Size & 128 \\
Rollout Number ($G$) & 8 \\
Training Steps & 500 / 300 \\
\midrule
Math Data & DeepMath-103K (L6) \\
Code Data & Eurus \\
\bottomrule
\end{tabular}

\end{minipage}
\end{table}

\begin{table}[t]
\centering
\caption{On-policy rollout prompts used during OPD training.}
\vspace{5pt}
\label{tab:train_prompt}
\resizebox{\columnwidth}{!}{%
\begin{tabular}{p{0.16\columnwidth} p{0.88\columnwidth}}
\toprule
\textbf{Domain} & \textbf{Prompt} \\
\midrule
Math &
\texttt{\{problem\}\textbackslash n Please reason step by step, and put your final answer within \textbackslash boxed\{\}.} \\
\addlinespace
Code &
\texttt{\{problem\}\textbackslash n Write Python code to solve the problem. Present the code in \textbackslash n```python\textbackslash n Your code\textbackslash n```\textbackslash n at the end. You need to think first then write the Python code.} \\
\bottomrule
\end{tabular}%
}

\end{table}

\section{Additional Experimental Results}
\subsection{Runtime Analysis}
\begin{wraptable}{r}{0.45\textwidth}
\vspace{-20pt}
\caption{Average wall-clock time per training step (s) in the pre-RL / post-RL setting; \ours adds only 20\% over OPD.}
\label{tab:efficiency}
\vspace{5pt}
\centering
\begin{tabular}{lc}
\toprule
\textbf{Method} & \textbf{Training time per step (s)}  \\
\midrule
OPD & 868\\
\ours & 1043\\
\bottomrule
\end{tabular}

\vspace{-5pt}
\end{wraptable}
Training efficiency matters for practical deployment. To quantify the overhead \ours introduces, we record the average wall-clock time per training step in the pre-RL / post-RL setting and report it in Table~\ref{tab:efficiency}. \ours incurs only a 20\% increase in per-step time over OPD, despite forwarding 3 frozen models (the 8B anchor and the 4B contrast pair) instead of a single 4B teacher. The overhead stays modest because the per-step cost is dominated by on-policy rollout generation, which the two methods share, while the additional teacher-side scoring is forward-only. These scoring passes are also independent of one another and can be further parallelized across devices.

\begin{wraptable}{r}{0.5\textwidth}
\centering
\vspace{-20pt}
\caption{Average math reasoning accuracy of \ours with base-model contrast pairs of different capability gaps. The student is Qwen3-8B.}
\vspace{5pt}
\label{tab:base_scale}
\begin{tabular}{lcc}
\toprule
\textbf{Contrast Pair} & 4B$-$0.6B & 4B$-$1.7B \\
\midrule
\rowcolor{oursbg}
\ours & 23.0 & 21.7 \\
\textit{Improv.\ over Student} & \imp{6.0} & \imp{4.7} \\
\bottomrule
\end{tabular}
\end{wraptable}
\subsection{Performance with Different Scales of Base Models}
\label{sec:base_scale}
To investigate how the choice of base models affects the Smaller / Larger setting, we fix the Qwen3-8B student and the Qwen3-4B positive model, and vary the negative model between Qwen3-0.6B and Qwen3-1.7B. As shown in Table~\ref{tab:base_scale}, both pairs improve the student, and the gain grows with the capability gap within the pair: the wider 4B$-$0.6B pair yields the larger improvement of 6.0 \% on math reasoning, whereas the narrower 4B$-$1.7B pair yields 4.7. We attribute this to the amount of capability the subtraction isolates, where a larger gap spans a longer stretch of the direction along which capability grows with scale, so the difference extracts a stronger and more informative signal, and re-anchoring it onto the student base provides supervision that pushes the student further. These results suggest instantiating \ours with the most separated base-model pair available.

\section{Example Generations}
\label{app:examples}
We provide qualitative examples shown in Figures~\ref{fig:example_math}--\ref{fig:example_code_scale}. Figures~\ref{fig:example_math} and~\ref{fig:example_code} come from the pre-RL / post-RL setting and compare \ours against OPD; Figures~\ref{fig:example_math_scale} and~\ref{fig:example_code_scale} come from the smaller/larger setting and compare the Qwen3-8B student before and after \ours distillation. In each case the \ours student succeeds while the baseline fails, with the baseline's errors marked in red.
\newtcolorbox{ReasoningBox}[1][]{
  enhanced, colback=gray!3, colframe=black!65, boxrule=0.6pt,
  left=2mm, right=2mm, top=1.5mm, bottom=1.5mm,
  fonttitle=\bfseries\small, fontupper=\small, #1}
\newcommand{\opdwrong}[1]{\textcolor{red!80!black}{#1}}

\begin{ReasoningBox}[title={Case Study: Math Reasoning (Pre-RL / Post-RL Setting)}, breakable]
\textbf{Question:} Let $p$ be the least prime number for which there exists a positive integer $n$ such that $n^{4}+1$ is divisible by $p^{2}$. Find the least positive integer $m$ such that $m^{4}+1$ is divisible by $p^{2}$.\\
\textbf{Correct Answer:} \textcolor{forestgreen}{110}

\medskip
\noindent\rule{\linewidth}{0.4pt}
\medskip

\textbf{Direct OPD:}

We need to find the least prime $p$ such that there exists a positive integer $n$ with $n^4+1$ divisible by $p^2$. Then, once we find that $p$, we need to find the smallest positive integer $m$ such that $m^4+1$ is divisible by $p^2$.

\textbf{Step 1: Understand the problem.} We are looking for the smallest prime $p$ such that there exists an integer $n$ where $n^4 + 1 \equiv 0 \pmod{p^2}$, \ie $n^4 \equiv -1 \pmod{p^2}$. Then, once we find such $p$, we need to find the minimal positive integer $m$ satisfying the same congruence.

\textbf{Step 2: Find the minimal prime $p$.} To solve this congruence, we need to know when $-1$ is a quartic residue. Let's try small primes:

\textit{Try $p=2$:} we need $n^4+1$ divisible by 4. $n=1$: $2$; $n=2$: $17$; $n=3$: $82$; $n=4$: $257 \equiv 1 \pmod 4$ --- none divisible by 4. No solution.

\textit{Try $p=3$:} we need $n^4 \equiv -1 \equiv 8 \pmod 9$. Checking $n = 0, \dots, 8$: $0, 1, 7, 0, 4, 4, 0, 7, 1$ (\eg $5^2 = 25 \equiv 7$, so $5^4 \equiv 7^2 = 49 \equiv 4$; $8 \equiv -1$, so $8^4 \equiv 1$). None give 8. No solution.

\textit{Try $p=5$:} check mod 5 first: $x^4 \bmod 5$ for $x = 0,\dots,4$ gives $0, 1, 1, 1, 1$, so all $x^4 \equiv 0$ or $1 \pmod 5$, and $-1 \equiv 4$ is not a quartic residue. Not possible.

\textit{Try $p=7$:} $x^4 \bmod 7$ takes values $\{0, 1, 2, 4\}$ ($2^4 = 16 \equiv 2$; $3^4 = 81 \equiv 4$; $5^4 \equiv 2$; $6 \equiv -1$ gives 1). $-1 \equiv 6$ is not among them. No solution.

\textit{Try $p=11$:} computing $x^2$ then $x^4$ mod 11 for $x = 1, \dots, 10$ gives quartic residues $\{1, 3, 4, 5, 9\}$; $-1 \equiv 10$ is not among them. No solution.

\textit{Try $p=13$:} $-1 \equiv 12$. Checking: $3^4 \equiv 3$; $4^4 \equiv 9$; $5^2 = 25 \equiv 12$, $5^4 \equiv 12^2 = 144 \equiv 1$; $6^4 \equiv 9$; $7^4 \equiv 9$; $8^2 = 64 \equiv 12$, $8^4 \equiv 1$; $9^4 \equiv 9$; $10^4 \equiv 3$; $11^4 \equiv 3$; $12^4 \equiv 1$. No $x$ with $x^4 \equiv 12$. No solution.

Maybe we can think about orders: in the cyclic group mod $p$, the equation $x^4 \equiv -1$ has solutions iff $-1$ lies in the image of the 4th-power map\dots this is getting complex. Let me try concrete examples.

\textit{Try $p=17$:} try $x = 2$: $2^4 = 16 \equiv -1 \pmod{17}$. Yes! So $2^4 \equiv -1 \pmod{17}$, and $p = 17$ is such a prime. Since we checked $p = 2, 3, 5, 7, 11, 13$ and found no solutions, and there is no prime between 13 and 17, $p = 17$ is the smallest such prime.

\textbf{Step 3: Find the least $m$ with $m^4 + 1 \equiv 0 \pmod{289}$.} We know $2^4 \equiv -1 \pmod{17}$, but $2^4 + 1 = 17$ is divisible by 17, not by $17^2$. So we lift with Hensel's lemma. Write $m = 2 + 17k$; modulo $17^2$, terms with $(17k)^2$ vanish, so
\[
(2 + 17k)^4 \equiv 2^4 + 4 \cdot 2^3 \cdot (17k) = 16 + 32 \cdot 17k \pmod{289}.
\]
We want $16 + 544k \equiv -1 \pmod{289}$. Since $544 \equiv 255 \pmod{289}$, this is $255k \equiv -17 \pmod{289}$, and $255 \equiv -34$, so $-34k \equiv -17$, \ie $34k \equiv 17 \pmod{289}$. Dividing by 17: $2k \equiv 1 \pmod{17}$, so $k \equiv 9 \pmod{17}$ (since $2 \times 9 = 18 \equiv 1$). Therefore $m = 2 + 17 \times 9 = 155$.

Is there a smaller $m$? \opdwrong{If we consider $m \equiv 2 \pmod{17}$, the possible $m$'s are $2, 19, 36, 53, 70, 87, 104, 121, 138, 155, \dots$} Check $m = 2$: $2^4 + 1 = 17$, not divisible by 289. Check $m = 19$: $19^2 = 361 \equiv 72 \pmod{289}$, $19^4 \equiv 72^2 = 5184 \equiv 271 \pmod{289}$ (since $289 \times 17 = 4913$ and $5184 - 4913 = 271$), so $19^4 + 1 \equiv 272 \neq 0$. Not divisible. Checking $m = 36$ would be tedious. \opdwrong{Alternatively, note that the solution is unique in the congruence class, so the minimal $m$ is 155.} Since we used Hensel's lemma and the lifting worked, we can trust that this is the correct answer. Thus, the least positive integer $m$ is 155.

\opdwrong{\textbf{Final Answer:} $\boxed{155}$} \quad \opdwrong{($\leftarrow$ the lift itself is valid, $155^4 \equiv -1 \bmod 289$, but the trace only examines the root $m \equiv 2 \pmod{17}$; the congruence $x^4 \equiv -1 \pmod{17}$ has 4 roots ($2, 8, 9, 15$), and lifting the root $8$ gives $110 < 155$.)}

\medskip
\noindent\rule{\linewidth}{0.4pt}
\medskip

\textbf{\ours:}

We are given a problem involving prime numbers and divisibility. Let me break it down step by step.

\textbf{Step 1: Understand the problem.} We need two things: (1) find the smallest prime $p$ such that $p^2$ divides $n^4+1$ for some positive integer $n$; (2) for that $p$, find the smallest $m$ such that $p^2$ divides $m^4+1$. So first, we need to determine what primes $p$ satisfy the condition that there exists some $n$ where $n^4 \equiv -1 \pmod{p^2}$.

\textbf{Step 2: Analyze the equation $n^4 \equiv -1 \pmod{p^2}$.} This is equivalent to $n^8 \equiv 1 \pmod{p^2}$, since squaring both sides gives $(n^4)^2 \equiv (-1)^2 = 1 \pmod{p^2}$. So the order of $n$ modulo $p^2$ must divide 8, but not divide 4 (since $n^4 \equiv -1 \neq 1$). Let me start checking small primes and see which ones satisfy this condition.

\textbf{Step 3: Check small primes.}
For $p=2$: $n=1$ gives $1+1=2$, not divisible by $2^2=4$; $n=2$ gives $16+1=17$, not divisible by 4. No solution.
For $p=3$: $n=1,\dots,6$ give $n^4+1 \equiv 2, 8, 1, 5, 5, 1 \pmod 9$, none zero. No solution.
For $p=5$: checking $n=1,\dots,10$ against $p^2=25$, the values $n^4+1 \bmod 25$ are $2, 17, 7, 7, 1, 22, 2, 22, 12, 1$, none zero. No solution.

For larger primes, let me think algebraically. If $p$ divides $n^4+1$, then $n^4 \equiv -1 \pmod p$, hence $n^8 \equiv 1 \pmod p$, so the order of $n$ modulo $p$ is exactly 8. Since the multiplicative group mod $p$ is cyclic of order $p-1$, we need $8 \mid p-1$, \ie $p \equiv 1 \pmod 8$. This rules out $p = 7, 13$ (verified directly: no $n$ with $n^4 \equiv -1$ exists for either). The smallest prime with $p \equiv 1 \pmod 8$ is $p=17$.

\textbf{Step 4: Verify $p=17$.} The group mod 17 has order 16, so elements of order 8 exist. Computing $n^4 \bmod 17$ for small $n$: at $n=8$, $8^2 = 64 \equiv 13$, so $8^4 \equiv 13^2 = 169 \equiv -1 \pmod{17}$. So $n=8$ satisfies $n^4+1 \equiv 0 \pmod{17}$.

But we need $p^2 \mid n^4+1$. For $n=8$: $8^4+1 = 4097$, and $4097 = 14 \times 289 + 51$, so $4097 \equiv 51 \pmod{289}$ --- divisible by 17 but not by $17^2$. So we must lift the solution.

\textbf{Step 5: Hensel lifting.} Write $n = 8 + 17k$ and expand modulo $17^2$, dropping terms with $(17k)^2$:
\[
(8+17k)^4 + 1 \;\equiv\; 8^4 + 4 \cdot 8^3 \cdot 17k + 1 \pmod{289}.
\]
Since $8^4 = 4096 \equiv 50 \pmod{289}$, the condition becomes $51 + 4 \cdot 8^3 \cdot 17 k \equiv 0 \pmod{289}$. Note $51 = 3 \times 17$, so dividing through by 17 reduces the condition to
\[
4 \cdot 8^3 \cdot k \;\equiv\; -3 \pmod{17}.
\]
Compute $8^2 \equiv 13$, $8^3 \equiv 8 \cdot 13 = 104 \equiv 2 \pmod{17}$, so $4 \cdot 8^3 \equiv 8 \pmod{17}$ and the condition is $8k \equiv -3 \equiv 14 \pmod{17}$. The inverse of 8 mod 17 is 15 (since $8 \times 15 = 120 \equiv 1$), giving $k \equiv 14 \times 15 = 210 \equiv 6 \pmod{17}$. Therefore $n = 8 + 17 \times 6 = 110$ satisfies $n^4 + 1 \equiv 0 \pmod{289}$.

\textbf{Step 6: Check that no smaller $m$ works.} Any solution must satisfy $m^4 \equiv -1 \pmod{17}$, so $m \equiv 8$ or $m \equiv 9 \pmod{17}$ (among the order-8 residues). For the root $m \equiv 9 \pmod{17}$: $9^4 = 6561 \equiv 203 \pmod{289}$, and repeating the lifting with $9^3 \equiv 15$, $4 \cdot 9^3 \equiv 9 \pmod{17}$ gives $9k \equiv -12 \equiv 5 \pmod{17}$; the inverse of 9 is 2, so $k \equiv 10 \pmod{17}$ and the minimal solution on this branch is $m = 9 + 170 = 179 > 110$.

For the branch $m \equiv 8 \pmod{17}$, the candidates below 110 are $8, 25, 42, 59, 76, 93$, corresponding to $k = 0, 1, \dots, 5$; since the lifting requires $k \equiv 6 \pmod{17}$, none of them works (checked explicitly for $m = 8$: $4097 \equiv 51$; $m = 25$: $25^2 = 625 \equiv 47$, $47^2 = 2209 \equiv 186$, $186 + 1 = 187 \neq 0 \pmod{289}$). Therefore $m = 110$ is the least positive integer with $17^2 \mid m^4+1$.

\textbf{Final Answer:} $\boxed{110}$
\end{ReasoningBox}
\captionof{figure}{Math case study under Pre-RL/ Post-RL setting, with the errors of direct OPD marked in red.}
\label{fig:example_math}

\bigskip

\begin{ReasoningBox}[title={Case Study: Code Generation (Pre-RL/Post-RL Setting)}, breakable]
\textbf{Question:} Write a Python function to find the last digit of a given number.

\medskip
\noindent\rule{\linewidth}{0.4pt}
\medskip

\textbf{Direct OPD:}

To find the \textbf{last digit} of a given number, we can use the \textbf{modulo 10} operation.

\textbf{Reasoning:} For any number \texttt{n}, the last digit is the remainder when \texttt{n} is divided by 10. For example: \texttt{123 \% 10 = 3} $\to$ last digit is 3; \texttt{456 \% 10 = 6} $\to$ last digit is 6; \texttt{0 \% 10 = 0} $\to$ last digit is 0; \texttt{-123 \% 10 = 7} $\to$ because $-123$ is equivalent to $-123 + 130 = 7$ (in mod 10).

But wait! When the number is negative, the modulo operation in Python returns a non-negative result. \opdwrong{So for negative numbers, the last digit is still the correct one.}

So, the function can be written as:

\hspace*{1em}\texttt{def last\_Digit(n):}\\
\hspace*{2.5em}\opdwrong{\texttt{return n \% 10}}

\textbf{Test Case:} \texttt{last\_Digit(123)} $\to$ \texttt{3}; \texttt{last\_Digit(456)} $\to$ \texttt{6}; \opdwrong{\texttt{last\_Digit(-123)} $\to$ \texttt{7} (because \texttt{-123 \% 10} is \texttt{7})}; \texttt{last\_Digit(0)} $\to$ \texttt{0}. So this function works for all integer inputs.

\textbf{Final Code:}\\
\hspace*{1em}\texttt{def last\_Digit(n):}\\
\hspace*{2.5em}\opdwrong{\texttt{return n \% 10}}\\[2pt]
\hspace*{1em}\texttt{\# Test the function}\\
\hspace*{1em}\texttt{assert last\_Digit(123) == 3}

\opdwrong{($\leftarrow$ the intended last digit of $-123$ is $3$, not $7$; the trace convinces itself that Python's non-negative modulo is already the desired behavior and never takes the absolute value.)}

\medskip
\noindent\rule{\linewidth}{0.4pt}
\medskip

\textbf{\ours:}

Let's first \textbf{think} about the problem:

\textbf{Problem:} We want a function that returns the \textbf{last digit} of a given number. The last digit of a number is simply the remainder when the number is divided by 10. For example: \texttt{123 \% 10} gives \texttt{3}; \texttt{4567 \% 10} gives \texttt{7}; \texttt{0 \% 10} gives \texttt{0}.

So, the strategy is straightforward: take the given number \texttt{n} and return \texttt{n \% 10}.

But we must consider the possibility of \textbf{negative numbers}. For instance: \texttt{-123 \% 10} gives \texttt{7} in Python --- this is not the intended behavior for this problem. We want the \textbf{absolute value's last digit}. So, we should take the absolute value of \texttt{n} before taking the modulo 10.

\textbf{So the correct formula is: last digit = abs(n) \% 10.}

\textbf{Let's test with sample input:}
\texttt{last\_Digit(123)} $\to$ \texttt{abs(123) \% 10} $\to$ \texttt{3} \ding{51};\;
\texttt{last\_Digit(-456)} $\to$ \texttt{abs(-456) \% 10} $\to$ \texttt{6} \ding{51};\;
\texttt{last\_Digit(0)} $\to$ \texttt{0 \% 10} $\to$ \texttt{0} \ding{51};\;
\texttt{last\_Digit(9)} $\to$ \texttt{9 \% 10} $\to$ \texttt{9} \ding{51}.

Now let's write the Python function.

\textbf{Python Code:}\\
\hspace*{1em}\texttt{def last\_Digit(n):}\\
\hspace*{2.5em}\texttt{return abs(n) \% 10}

\textbf{Now, test it with the assert statement:}\\
\hspace*{1em}\texttt{assert last\_Digit(123) == 3}\\
This should pass without error.

Let me present the final code as you requested.

\textbf{\ding{51} Final Answer:}\\
\hspace*{1em}\texttt{def last\_Digit(n):}\\
\hspace*{2.5em}\texttt{return abs(n) \% 10}
\end{ReasoningBox}
\captionof{figure}{Code case study under the pre-RL/post-RL setting with the errors of direct OPD marked in red.}
\label{fig:example_code}

\bigskip

\begin{ReasoningBox}[title={Case Study: Math Reasoning (Smaller \& Larger Setting)}, breakable]
\textbf{Question:} Let $ABCD$ and $CEFG$ be squares such that $C$ lies on segment $\overline{DG}$ and $E$ lies on segment $\overline{BC}$. Let $O$ be the circumcenter of triangle $AEG$. Given that $A, D,$ and $O$ are collinear and $AB = 1$, compute $FG$.\\
\textbf{Correct Answer:} \textcolor{forestgreen}{$\sqrt{3}-1$}

\medskip
\noindent\rule{\linewidth}{0.4pt}
\medskip

\textbf{Qwen3-8B (student):}

We are given two squares, $ABCD$ and $CEFG$, such that $AB = 1$; $C$ lies on segment $\overline{DG}$; $E$ lies on segment $\overline{BC}$; $O$ is the \textbf{circumcenter} of triangle $AEG$; and $A, D,$ and $O$ are \textbf{collinear}. We are to find the length of $FG$.

\textbf{Step 1: Assign coordinates.} Place square $ABCD$ with $A = (0,0)$, $B = (1,0)$, $C = (1,1)$, $D = (0,1)$, so square $ABCD$ has side length 1. Since $E$ lies on $\overline{BC}$, a vertical segment from $(1,1)$ to $(1,0)$, let $E = (1, t)$ with $0 \le t \le 1$. The direction from $C$ to $E$ is downward, so the square $CEFG$ extends down and left from point $C$. The vector from $C$ to $E$ is $(0, t-1)$, and the vector from $E$ to $F$ should be orthogonal of the same length, \opdwrong{so rotate the vector $(0, t-1)$ $90^\circ$ clockwise: the rotation of $(x,y)$ clockwise is $(y, -x)$, so the vector from $E$ to $F$ is $(t-1, 0)$, giving $F = (1 + t - 1,\, t) = (t, t)$.} Then from $F$, move by the same vector from $C$ to $E$: $G = F + (0, t-1) = (t,\, 2t-1)$. Therefore the square $CEFG$ has vertices $C = (1,1)$, $E = (1,t)$, \opdwrong{$F = (t, t)$, $G = (t, 2t-1)$}.

\textbf{Step 2: Triangle $AEG$.} We have $A = (0,0)$, $E = (1,t)$, $G = (t, 2t-1)$. We will find the circumcenter $O$ and then enforce that $A$, $D$, $O$ are collinear.

\textbf{Step 3: Find the circumcenter of triangle $AEG$.} Using perpendicular bisectors. For side $AE$: midpoint $M_1 = (\tfrac12, \tfrac t2)$, slope $t$, so the perpendicular bisector is
\[
y = -\frac{1}{t}x + \frac{1 + t^2}{2t}. \tag{1}
\]
For side $AG$: midpoint $M_2 = (\tfrac t2, \tfrac{2t-1}{2})$, slope $\tfrac{2t-1}{t}$, perpendicular slope $-\tfrac{t}{2t-1}$, so
\[
y = -\frac{t}{2t-1}x + \frac{t^2}{2t-1} + \frac{2t-1}{2} = -\frac{t}{2t-1}x + \frac{6t^2 - 4t + 1}{2(2t-1)}. \tag{2}
\]
Solving (1) and (2) directly is complicated, but we will use a key condition: $A$, $D$, and $O$ are collinear. Since $A = (0,0)$ and $D = (0,1)$, line $AD$ is vertical, \ie $x = 0$. So the $x$-coordinate of $O$ must be 0, and from (1): $O = \big(0, \frac{1+t^2}{2t}\big)$.

\textbf{Step 4: Enforce that $O$ lies on both perpendicular bisectors.} Plugging $x = 0$ into (2) gives $y = \frac{6t^2 - 4t + 1}{2(2t-1)}$, so for consistency we require
\[
\frac{1 + t^2}{2t} = \frac{6t^2 - 4t + 1}{2(2t-1)}
\;\Longrightarrow\;
(1 + t^2)(2t - 1) = t(6t^2 - 4t + 1).
\]
Expanding: $2t^3 - t^2 + 2t - 1 = 6t^3 - 4t^2 + t$, which gives the cubic
\[
-4t^3 + 3t^2 + t - 1 = 0.
\]
Try rational roots. $t = 1$: $-4 + 3 + 1 - 1 = -1 \neq 0$. $t = \tfrac12$: $-\tfrac12 + \tfrac34 + \tfrac12 - 1 = -\tfrac14 \neq 0$. $t = \tfrac14$: $-\tfrac{1}{16} + \tfrac{3}{16} + \tfrac{4}{16} - \tfrac{16}{16} = -\tfrac{11}{16} \neq 0$. $t = \tfrac23$: messy. Let's try an interval method: $t = 1$ gives $-1$; $t = \tfrac32$: messy, but try a better approach.

\opdwrong{Wait, this is very algebra-heavy. But we already found that $O = \big(0, \frac{1+t^2}{2t}\big)$, and from the earlier step, when $t = \tfrac12$, this works? Plugging $t = \tfrac12$: $\frac{1 + 1/4}{1} = \tfrac54$, the $y$-coordinate of $O$; also, from tentative earlier steps, this works. So let's assume $t = \tfrac12$.}

\textbf{Step 5: Compute $FG$ with $t = \tfrac12$.} Recall $F = (t,t) = (\tfrac12, \tfrac12)$ and $G = (t, 2t-1) = (\tfrac12, 0)$. So segment $FG$ is vertical with length $|\tfrac12 - 0| = \tfrac12$.

\opdwrong{\textbf{Final Answer:} $\boxed{\tfrac12}$} \quad \opdwrong{($\leftarrow$ two failures: the square is built by rotating clockwise, placing $G$ at $(t, 2t-1)$, for which $C$ never lies on $\overline{DG}$, and this constraint is never checked; then, unable to solve the resulting cubic, the trace assumes $t = \tfrac12$ immediately after computing that $t = \tfrac12$ does not satisfy the equation.)}

\medskip
\noindent\rule{\linewidth}{0.4pt}
\medskip

\textbf{\ours:}

We are given two squares: $ABCD$ and $CEFG$, with the following constraints: point $C$ lies on segment $\overline{DG}$; point $E$ lies on segment $\overline{BC}$; $O$ is the \textbf{circumcenter} of triangle $AEG$; points $A$, $D$, and $O$ are \textbf{collinear}; $AB = 1$. We are to compute the length $FG$. We'll solve this \textbf{step by step} using coordinate geometry.

\textbf{Step 1: Set up a coordinate system.} Let's place square $ABCD$ in the coordinate plane for simplicity. Since $AB = 1$ and $ABCD$ is a square, we can assign $A = (0, 0)$, $B = (1, 0)$, $C = (1, 1)$, $D = (0, 1)$. So square $ABCD$ is in the first quadrant, with sides aligned with the axes.

\textbf{Step 2: Place square $CEFG$.} We need to place square $CEFG$ such that $C = (1,1)$, $E$ lies on $\overline{BC}$, $C$ lies on $\overline{DG}$, and $O$ is the circumcenter of $AEG$ with $A$, $D$, $O$ collinear. Let's try to find the coordinates of $E$, $F$, and $G$.

\textbf{Step 3: Place point $E$ on $\overline{BC}$.} Segment $\overline{BC}$ goes from $B = (1,0)$ to $C = (1,1)$, so it is a \textbf{vertical segment} at $x = 1$, from $y = 0$ to $y = 1$. Let's say $E = (1, e)$, where $0 < e < 1$.

\textbf{Step 4: Place square $CEFG$.} Side $\overrightarrow{CE}$ is \textbf{vertical}, going from $C = (1,1)$ to $E = (1,e)$, so it is of length $|1 - e|$. The square can be constructed in two possible orientations, either to the left or to the right of $CE$, and we must choose the one consistent with the constraint that $C$ lies on $\overline{DG}$. Try \textbf{counterclockwise} first: $\vec{CE} = (0, e - 1)$; rotating $\vec{CE}$ $90^\circ$ counterclockwise gives $(1 - e, 0)$; so from $E = (1,e)$, moving in direction $(1-e, 0)$ gives $F = (2 - e,\, e)$; then from $F$, moving opposite to $\vec{CE}$ by $(0, 1-e)$ gives $G = (2 - e,\, 1)$.

Let's verify this shape is a square: $\vec{CE} = (0, e-1)$, $\vec{EF} = (1-e, 0)$, $\vec{FG} = (0, 1-e)$, $\vec{GC} = (e-1, 0)$. All sides are of length $|e-1|$, and adjacent sides are perpendicular $\to$ this is indeed a square.

Now, also note that segment $\overline{DG}$ goes from $D = (0,1)$ to $G = (2-e, 1)$, so $\overline{DG}$ is a \textbf{horizontal segment at $y = 1$}, from $x = 0$ to $x = 2-e$. So point $C = (1,1)$ lies on $\overline{DG}$ if and only if $0 \le 1 \le 2 - e \Rightarrow e \le 1$, which is true since $0 < e < 1$. \ding{51} So this configuration satisfies the condition that $C$ lies on segment $\overline{DG}$.

\textbf{Step 5: Coordinates of all relevant points.} Recall: $A = (0,0)$, $D = (0,1)$, $E = (1,e)$, $G = (2-e, 1)$. We need to find the \textbf{circumcenter $O$} of triangle $AEG$, and use the condition that $A$, $D$, and $O$ are collinear.

\textbf{Step 6: Find the circumcenter $O$ of triangle $AEG$.} Denote $O = (h,k)$. Since $O$ is the circumcenter, it is equidistant from all three vertices:
\[
h^2 + k^2 = (h - 1)^2 + (k - e)^2 = (h - (2 - e))^2 + (k - 1)^2.
\]

\textit{Equation 1: $OA^2 = OE^2$.} Expanding the right side: $(h-1)^2 + (k-e)^2 = h^2 - 2h + 1 + k^2 - 2ek + e^2$. Setting it equal to $h^2 + k^2$ and canceling $h^2, k^2$ from both sides:
\[
0 = -2h + 1 - 2ek + e^2 \;\Longrightarrow\; 2h + 2ek = 1 + e^2. \quad \text{(Equation A)}
\]

\textit{Equation 2: $OE^2 = OG^2$.} The left side is $h^2 - 2h + 1 + k^2 - 2ek + e^2$; the right side is $h^2 - 2(2-e)h + (2-e)^2 + k^2 - 2k + 1$. Canceling $h^2$, $k^2$, $1$ and bringing everything to the left:
\[
-2h - 2ek + e^2 + 4h - 2eh + 2k - (4 - 4e + e^2) = 0,
\]
which simplifies term by term ($-2h + 4h = 2h$; $-2ek - 2eh = -2e(k+h)$; $e^2 - e^2 = 0$) to
\[
2h - 2e(h + k) + 2k - 4 + 4e = 0.
\]
Dividing by 2 and grouping like terms: $(h + k)(1 - e) + 2(e - 1) = 0$. Since $e - 1 = -(1-e)$, this factors as
\[
(1 - e)\,[\,h + k - 2\,] = 0.
\]
So either $1 - e = 0 \Rightarrow e = 1$, \textbf{but} this is \textbf{not allowed} since then $E = C$, which would collapse the square; or $h + k = 2$. \quad \text{(Equation B)}

\textbf{Step 7: Combine equations (A) and (B).} From (B), $k = 2 - h$. Substituting into (A):
\[
2h + 2e(2 - h) = 1 + e^2 \;\Longrightarrow\; 2h + 4e - 2eh = 1 + e^2 \;\Longrightarrow\; 2h(1 - e) = 1 + e^2 - 4e,
\]
so
\[
h = \frac{1 + e^2 - 4e}{2(1 - e)}.
\]

\textbf{Step 8: Use collinearity of $A = (0,0)$, $D = (0,1)$, $O = (h,k)$.} Points $A$ and $D$ are on the \textbf{vertical line $x = 0$}, so for $O$ to be on line $AD$ it must also satisfy $h = 0$. Setting the numerator to zero:
\[
1 + e^2 - 4e = 0 \;\Longrightarrow\; e^2 - 4e + 1 = 0 \;\Longrightarrow\; e = \frac{4 \pm \sqrt{16 - 4}}{2} = 2 \pm \sqrt{3}.
\]
Now $0 < e < 1$, so check values: $2 + \sqrt{3} \approx 3.732$ is too big; $2 - \sqrt{3} \approx 0.2679$ is valid. So $e = 2 - \sqrt{3}$.

\textbf{Step 9: Find $FG$.} Recall $F = (2-e,\, e)$ and $G = (2-e,\, 1)$. Segment $FG$ is \textbf{vertical}, same $x$-coordinate, difference in $y$-coordinates:
\[
FG = |1 - e| = 1 - e = 1 - (2 - \sqrt{3}) = \sqrt{3} - 1.
\]

\ding{51} \textbf{Final Answer:} $\boxed{\sqrt{3} - 1}$
\end{ReasoningBox}
\captionof{figure}{Math case study from under the smaller/larger setting, with the errors of the Qwen3-8B student marked in red.}
\label{fig:example_math_scale}

\bigskip

\begin{ReasoningBox}[title={Case Study: Code Generation (Smaller \& Larger Setting)}, breakable]
\textbf{Question:} Given two lists \texttt{operator} and \texttt{operand}, where the first list contains basic algebra operations and the second is a list of integers, use the two lists to build an algebraic expression and return its evaluation, \eg \texttt{operator = ['+', '*', '-']} and \texttt{operand = [2, 3, 4, 5]} represent $2 + 3 \times 4 - 5 = 9$.

\medskip
\noindent\rule{\linewidth}{0.4pt}
\medskip

\textbf{Qwen3-8B (student):}

To solve this problem, we need to \opdwrong{apply the given list of operators to the corresponding elements of the operand list in a sequential manner}. Given that the length of the \texttt{operator} list is one less than the length of the \texttt{operand} list, we can construct the expression step-by-step and evaluate it.

Here's the step-by-step plan:
1. \textbf{Initialize the result} with the first operand.\\
\opdwrong{2. \textbf{Iterate} over the operators and operands in pairs.}\\
\opdwrong{3. \textbf{Apply each operator} to the current result and the next operand.}\\
4. \textbf{Return the final result} after all operations are completed.

Here is the Python code that implements the above logic:

\hspace*{1em}\texttt{def do\_algebra(operator, operand):}\\
\hspace*{2.5em}\texttt{result = operand[0]}\\
\hspace*{2.5em}\texttt{for i in range(len(operator)):}\\
\hspace*{4em}\texttt{op = operator[i]}\\
\hspace*{4em}\texttt{num = operand[i + 1]}\\
\hspace*{4em}\opdwrong{\texttt{if op == '+': result += num}}\\
\hspace*{4em}\opdwrong{\texttt{elif op == '-': result -= num}}\\
\hspace*{4em}\opdwrong{\texttt{elif op == '*': result *= num}}\\
\hspace*{4em}\opdwrong{\texttt{elif op == '//': result //= num}}\\
\hspace*{4em}\opdwrong{\texttt{elif op == '**': result **= num}}\\
\hspace*{2.5em}\texttt{return result}

\textbf{Example Usage:}\\
\hspace*{1em}\opdwrong{\texttt{print(do\_algebra(['+', '*', '-'], [2, 3, 4, 5]))  \# Output: 9}}

This code handles all the required operations and follows the constraints specified in the problem statement.

\opdwrong{($\leftarrow$ evaluates strictly left to right, ignoring operator precedence: for \texttt{['+', '*', '-']} and \texttt{[2, 3, 4, 5]} the code computes $((2+3)\times 4)-5 = 15$, not the claimed 9 ($= 2 + 3\times 4 - 5$), so even its own example comment is inconsistent with the code.)}

\medskip
\noindent\rule{\linewidth}{0.4pt}
\medskip

\textbf{\ours:}

To solve this problem, we need to construct an algebraic expression using the given \texttt{operator} and \texttt{operand} lists, then evaluate it.

\textbf{Plan:}
1. The \texttt{operator} list has one fewer element than the \texttt{operand} list.\\
2. We can iterate through the \texttt{operator} list and apply each operator to consecutive elements from the \texttt{operand} list.\\
3. Use Python's \texttt{eval()} function to safely evaluate the constructed expression string.\\
4. Make sure the expression string is properly formatted with spaces to avoid any ambiguity or syntax errors.

\textbf{Important Notes:} We assume that the input is valid as per the problem constraints. We use \texttt{eval()} with caution, but in this context it's safe since the inputs are controlled and only contain allowed operations and integers. We format the expression string with spaces to ensure correct parsing.

\textbf{Python Code:}

\hspace*{1em}\texttt{def do\_algebra(operator, operand):}\\
\hspace*{2.5em}\texttt{\# Build the expression string}\\
\hspace*{2.5em}\texttt{expression = str(operand[0])}\\
\hspace*{2.5em}\texttt{for i in range(len(operator)):}\\
\hspace*{4em}\texttt{expression += f"\ \{operator[i]\}\ \{operand[i + 1]\}"}\\
\hspace*{2.5em}\texttt{\# Evaluate the expression}\\
\hspace*{2.5em}\texttt{return eval(expression)}

\textbf{Example Usage:}\\
\hspace*{1em}\texttt{operator = ['+', '*', '-']}\\
\hspace*{1em}\texttt{operand = [2, 3, 4, 5]}\\
\hspace*{1em}\texttt{print(do\_algebra(operator, operand))  \# Output: 9}

\textbf{Explanation:} The expression string becomes \texttt{"2 + 3 * 4 - 5"}. Evaluating this gives $2 + 3 \times 4 - 5 = 2 + 12 - 5 = 9$. This solution is concise, readable, and leverages Python's built-in capabilities to achieve the desired result.
\end{ReasoningBox}
\captionof{figure}{Code case study under the smaller/larger setting, with the errors of the Qwen3-8B student marked in red.}
\label{fig:example_code_scale}

\end{document}